\pdfoutput=1

\documentclass[11pt]{article}

\usepackage[final]{acl}

\usepackage{times}
\usepackage{latexsym}

\usepackage[T1]{fontenc}

\usepackage[utf8]{inputenc}

\usepackage{microtype}
\usepackage{hyperref}
\usepackage{inconsolata}
\usepackage{authblk}

\usepackage{graphicx}

\title{VGA: Vision GUI Assistant - Minimizing Hallucinations through Image-Centric Fine-Tuning}
\author{
 \textbf{Ziyang Meng},
 \textbf{Yu Dai},
 \textbf{Zezheng Gong},
 \authorcr
 \textbf{Shaoxiong Guo},
 \textbf{Minglong Tang},
 \textbf{Tongquan Wei}\thanks{Corresponding author}
 \authorcr
East China Normal University
 \authorcr
{qdmengziyang@gmail.com, daiyuu@126.com}}

\begin{document}
\maketitle


\begin{abstract}
Large Vision-Language Models (VLMs) have already been applied to the understanding of Graphical User Interfaces (GUIs) and have achieved notable results. However, existing VLMs often overly rely on internal text-based knowledge while neglecting visual inputs. This imbalance may lead models to produce answers that do not align with the visual content in GUI comprehension tasks. Such inaccuracies are termed as `hallucinations' where models generate incorrect or illogical responses upon visual verification against GUI elements. These errors result in misinterpretations and diminish the model's practical utility in applied settings. To address these issues, we introduce VGA, a fine-tuned model designed for comprehensive GUI understanding. Our model aims to balance attention image and text to enhance interpretation and reduce hallucinations. We construct a Vision Question Answering (VQA) dataset of 63.8k high-quality examples with our propose \textit{Referent Method}, focusing on response with visual content of images. We then design a two-stage fine-tuning method to enhance both the model's accuracy to extract information from image content and alignment with human intent. Experiments show that our approach enhances the model's ability to extract information from images and achieves state-of-the-art results in GUI understanding tasks. 
Our dataset and fine-tuning script are available at \href{https://github.com/Linziyang1999/VGA-visual-GUI-assistant}{https://github.com/Linziyang1999/VGA-visual-GUI-assistant}

\end{abstract}

\begin{table*}[h]
    \centering
    \begin{tabular}{p{2.5cm}p{5cm}p{5cm}}
        \hline
        \textbf{\fontsize{12}{14}\selectfont Hallucination in GUI comprehension} & & \\
        \hline
        & \includegraphics[width=0.15\textwidth]{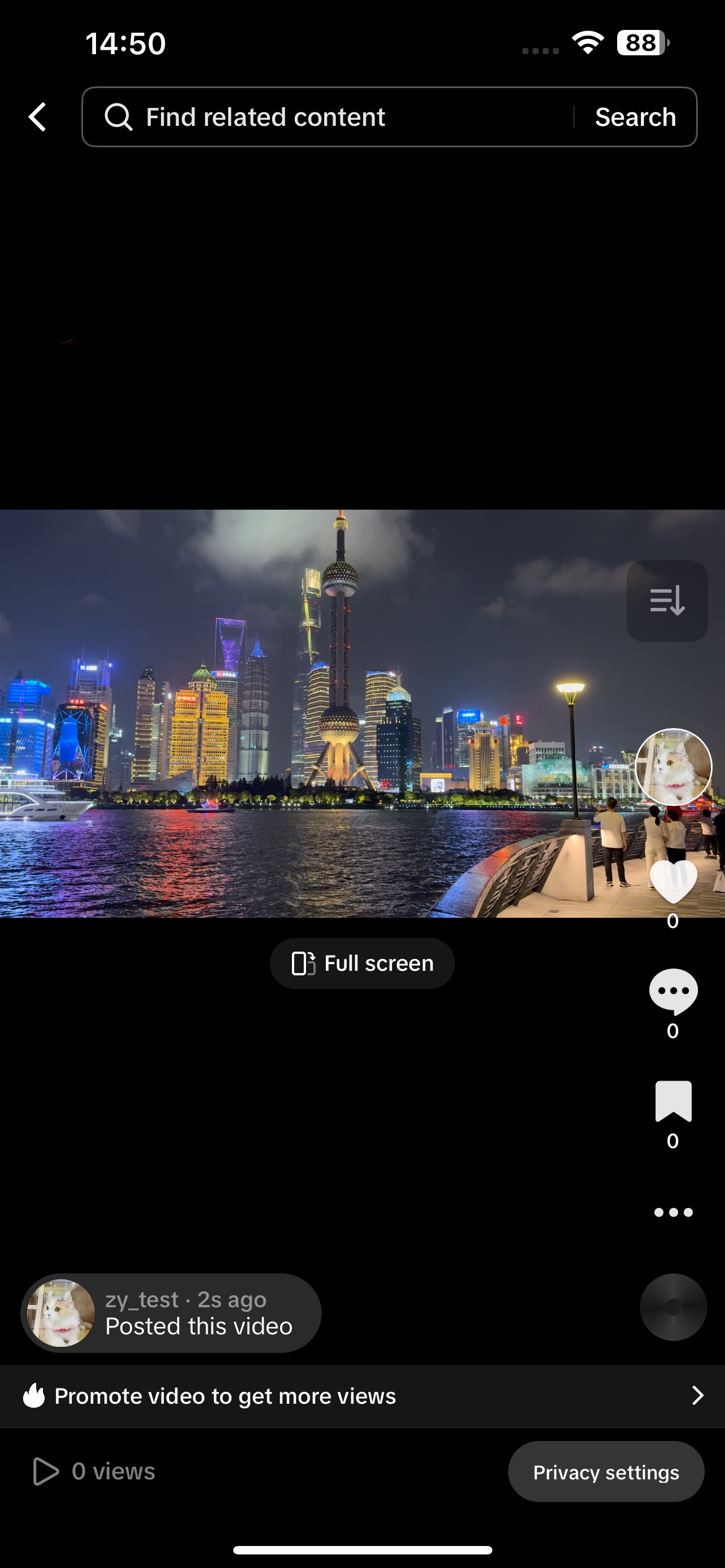} & \includegraphics[width=0.15\textwidth]{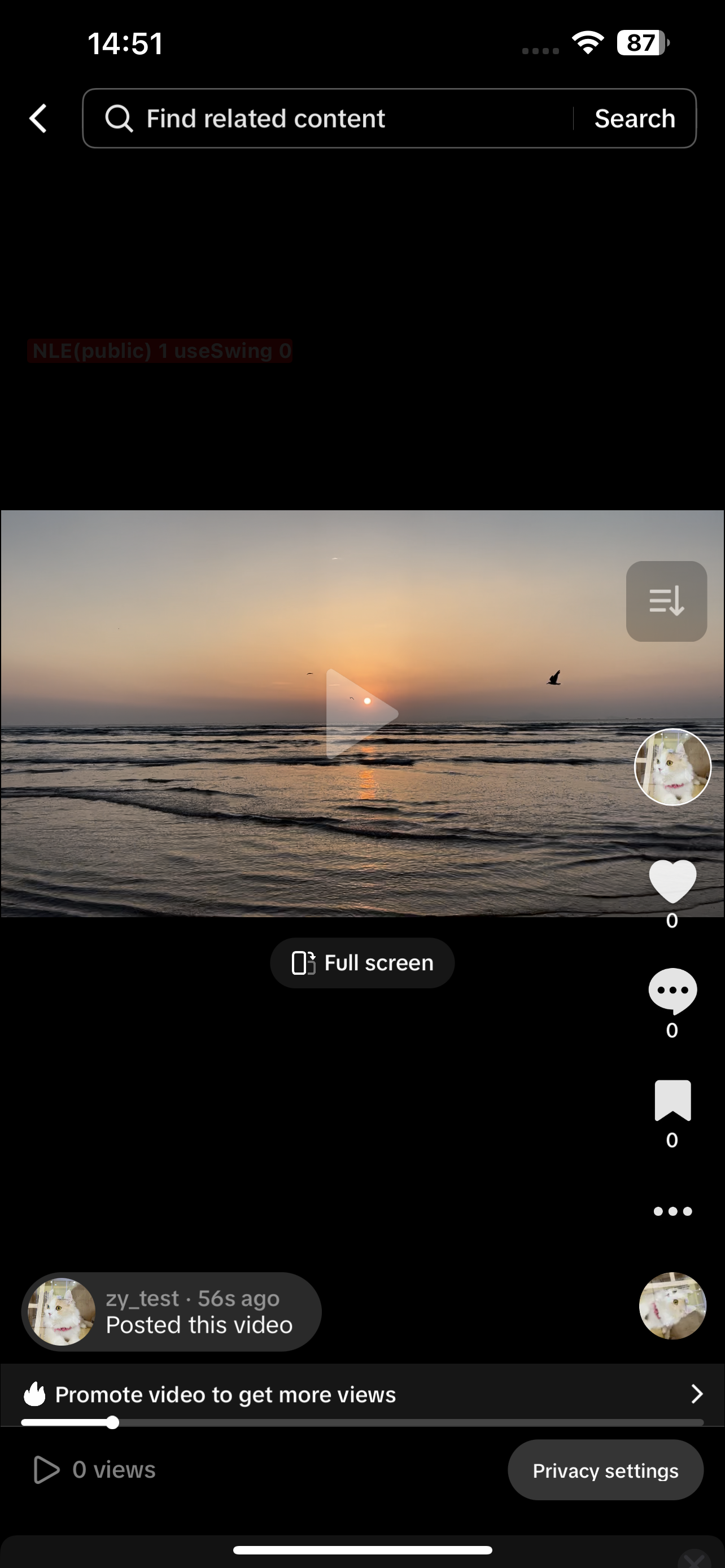}\\
        \fontsize{8}{10}\selectfont{User} & \fontsize{8}{10}\selectfont{\textcolor{blue}{How can I give a like to this video?}} & \fontsize{8}{10}\selectfont{\textcolor{blue}{How can I login?}} \\
        \fontsize{8}{10}\selectfont{LLaVA-v1.6-34B} & \fontsize{8}{10}\selectfont{Open the video you want to like. Look for a \textcolor{red}{thumbs-up or heart icon}, which is usually \textcolor{orange}{located near the bottom of the screen}...} & \fontsize{8}{10}\selectfont{Open the website or app of the social media platform you want to log in to. \textcolor{red}{Look for the "Login" or "Sign In" button}, which is usually \textcolor{orange}{located at the top right corner of the page}.} \\
        \hline
        \fontsize{8}{10}\selectfont{GPT-4V} & \fontsize{8}{10}\selectfont{...\textcolor{red}{there should be a heart icon or similar button} typically \textcolor{orange}{found below the video or in a sidebar}. Tapping on this icon...} & \fontsize{8}{10}\selectfont{Open the App or website... \newline \textcolor{red}{Find login section...} \newline Enter your credentials... } \\
        \hline
        \fontsize{8}{10}\selectfont{VGA} & \fontsize{8}{10}\selectfont{You can click the \textcolor{red}{heart-shaped icon} located on the \textcolor{orange}{right side of the screen}, just \textcolor{orange}{below the person's image}.} & \fontsize{8}{10}\selectfont{This page seems to be focused on displaying content and interacting with it, but \textcolor{red}{it doesn't appear to have a login function}.} \\
        \hline
    \end{tabular}
    \caption{Example of hallucination in GUI comprehension, red means element in image relate to answer and orange means location of the element which indicate if model really extract information from image.}
    \label{tab:hallucination}
    
\end{table*}

\section{Introduction}

Large Vision-Language Models (VLMs) have recently emerged as a powerful approach for various multimodal tasks. These models acquire textual knowledge through pre-training and develop image understanding abilities during instruction-tuning \cite{10.1145/3581783.3612498, liu2023improved}.
VLMs can effectively process both visual and linguistic information by using a visual-language projector, which maps different types of data into a shared latent space. For instance, LLaVA \cite{liu2024llavanext} and InstructBLIP \cite{dai2023instructblip} have demonstrated strong zero-shot capabilities in tasks like image captioning, visual reasoning, and complex conversations. UniChart \cite{masry2023unichart}, on the other hand, has demonstrated the powerful capabilities of VLMs in the understanding of formatted charts.


With the proliferation of mobile applications, the importance of Graphical User Interfaces (GUIs), which serve as a critical bridge between end users and applications, has increasingly garnered scholarly attention. GUIs, characterized by their structured layouts, rich graphical and textual content, and the inclusion of human operational logic \cite{banerjee2013graphical}, present a complex challenge: \textit{Can the success of VLMs be applied to the GUI domain?}

Traditional GUI comprehension method focus on conveying the actual user interface interactions and mirror the user's direct experience with the GUI. For instance, LabelDroid \cite{9284063} utilizes deep learning to predict labels for image-based buttons from a variety of commercial apps available on Google Play. Similarly, TANGO \cite{9402526} employs custom computer vision (CV) and text retrieval techniques to analyze visual and textual information on mobile screens.

However, traditional visual GUI comprehension methods focus mainly on identifying GUI components but fail to fully comprehend the graphical and textual information, layout, and the interaction context within the interface. In contrast, modern approaches such as those employed by ferret-UI and CogAgent, which utilize LVLMs, demonstrate significant advantages in this domain. These methods are designed to process both detailed textual and graphical content, leveraging their pre-trained capabilities. Nevertheless, both ferret-UI \cite{you2024ferretuigroundedmobileui} and CogAgent \cite{hong2023cogagentvisuallanguagemodel} frequently encounter limitations by over-relying on their pre-trained knowledge, which leads to a neglect of critical visual content. This often results in the generation of inaccurate or irrelevant responses, indicating a gap in their ability to achieve a balanced and holistic understanding of GUIs \cite{shahgir2024illusionvqa, zhang2023grounding}.
 
To address these issues, we propose VGA, a model fine-tuned on a self-construct 63.8K dataset, using a novel training method we design. During the dataset construction, we employ knowledge distillation from large language models to construct 63.8k dataset based on Rico \cite{10.1145/3126594.3126651}, and we adopt the \textit{Referent Method} to enhance the model's focus on image content by using visual and position information in constructing dataset, thereby increasing the relevance and accuracy of the responses. In the training process, we apply a Foundation and Advanced Comprehension (FAC) approach: the Foundation Stage enhances the model's understanding of GUI image, while the Advanced Comprehension stage improves the model's ability to respond to complex questions based on its understanding of the GUI. Additionally, we employ a task progression and reinforcement approach to create logical chains between tasks and within responses to strengthens the model's ability to reason and infer relationships within the provided context. Through these integrated strategies, VGA achieves state-of-the-art performance in GUI comprehension tasks.
Our primary contributions are as follows:



\begin{itemize}
    \item \textbf{Large-scale GUI corpus for LVLM Fine-tuning.} To fine-tune our model, we introduce a large-scale GUI corpus that includes a diverse array of apps accompanied by corresponding text descriptions and dialogues.
    \item \textbf{A Fine-tuned LVLM for GUI task.} We propose VGA, a LVLM for GUI comprehension, fine-tuned to fulfill both granular low-level and strategic high-level goals specialized for graphical user interfaces.
    \item \textbf{A Fine-tune Method to efficient improve LVLMs performance} We propose a two stage fine-tune method based on our dataset to achieve better understand of images.
    \item \textbf{Performance enhancements on real-world GUI tasks.}  As shown in experiment ~\ref{tab:bench}, we apply our VGA to a GUI comprehension bench, yielding promising results.
\end{itemize}



\section{Related Works}
\subsection{GUI Comprehension}

GUI comprehension is essential in mobile agent design to ensure app quality and user experience. It focuses on assessing interface components such as buttons, text boxes, and menus to verify their operational performance, visual design, and usability, enabling mobile agents to interact effectively with the app interface \cite{yu2023vision}. The software engineering community has been long-term focused on the improvement of mobile app GUI comprehension effectiveness in all aspects since last century \cite{10.1145/302405.302632}. The main purpose of these efforts is to understand GUI elements to advance GUI automation, including testing and other applications \cite{DBLP:journals/corr/abs-1812-11470, DBLP:journals/corr/abs-1801-06267, Said2020GUITF}.

Existing methods for GUI comprehension can be divided into code-based approaches, which rely on functional specifications or code analysis \cite{paiva2005modeling, el2010systematic}, and visual-based approaches, which improve performance by considering the GUI's visual representation through techniques like template matching and OCR \cite{cheng2019apply}. However, these methods still lack the ability to interact with GUIs in a human-like manner. Recently, Large Vision-Language Models (LVLMs) have emerged as a promising approach by combining visual and textual understanding. These models leverage vast prior knowledge gained from pre-train to comprehend human intentions and interact more naturally with GUI elements \cite{wang2024large, cui2024large}.

In our work, we leverage the strengths of LVLMs and enhance their performance in GUI comprehension using our constructed dataset and our designed training methods.

\subsection{Large Vision-Language Models}

The introduction of the transformer architecture has revolutionized natural language processing, enabling models to efficiently capture long-range dependencies and contextual information. This advance laid the groundwork for pre-trained Large Language Models (LLMs). The pre-train, fine-tune, and predict paradigm \cite{liu2021pretrain}, exemplified by models such as GPT \cite{radford2018improving} and BERT \cite{devlin2018bert}, led to significant improvements in language understanding and generation \cite{cheng2023adapting, yao2023empowering, arefeen2024leancontext, schick2020exploiting, 10366647}. Subsequent models like GPT-4 \cite{achiam2023gpt}, Llama \cite{touvron2023llama}, and Qwen \cite{bai2023qwen} have further expanded these capabilities. Explorations into the Mixture of Experts (MoE) architecture \cite{shen2023mixture, jiang2024mixtral} continue to enhance the scalability of Transformer-based models.

Large Vision-Language Models (LVLMs) harness the strengths of both LLMs and visual feature encoders, utilizing various methods to project visual data into an LLM-comprehensible space \cite{popescu2009multilayer,liu2023improved,liu2024visual,alayrac2022flamingo,wang2023cogvlm}. Models like Flamingo \cite{alayrac2022flamingo} and CogVLM \cite{wang2023cogvlm} exemplify the architectures that achieve this integration. Additionally, UniChart \cite{masry2023unichart} excels in chart comprehension by leveraging LVLMs to accurately interpret formatted charts and complex informational structures. Similarly, LLaVA \cite{liu2023improved} integrates pre-trained visual and language models to enhance the understanding of multimodal inputs, achieving superior performance in visual and textual comprehension tasks. 

In the field of graphical user interface (GUI) analysis, traditional methods like Ferret-UI \cite{you2024ferretuigroundedmobileui} and CogAgent \cite{hong2023cogagentvisuallanguagemodel} predominantly focus on text, largely because their Large Language Model (LLM) components have been thoroughly trained to emphasize textual information. This emphasis can lead to the visual components of the input being overlooked or misinterpreted in the responses, often resulting in the generation of irrelevant answers—a phenomenon known as "hallucination".

Inspired by these work, we investigate the performance of LVLMs in the context of complex GUI interfaces. Our study focuses on evaluating how well these models understand and interact with GUIs that feature structured layouts and rich interactions, aiming to further enhance their applicability and accuracy in real-world scenarios.

\begin{figure*}[h]
    \centering
    \includegraphics[width=0.9\textwidth]{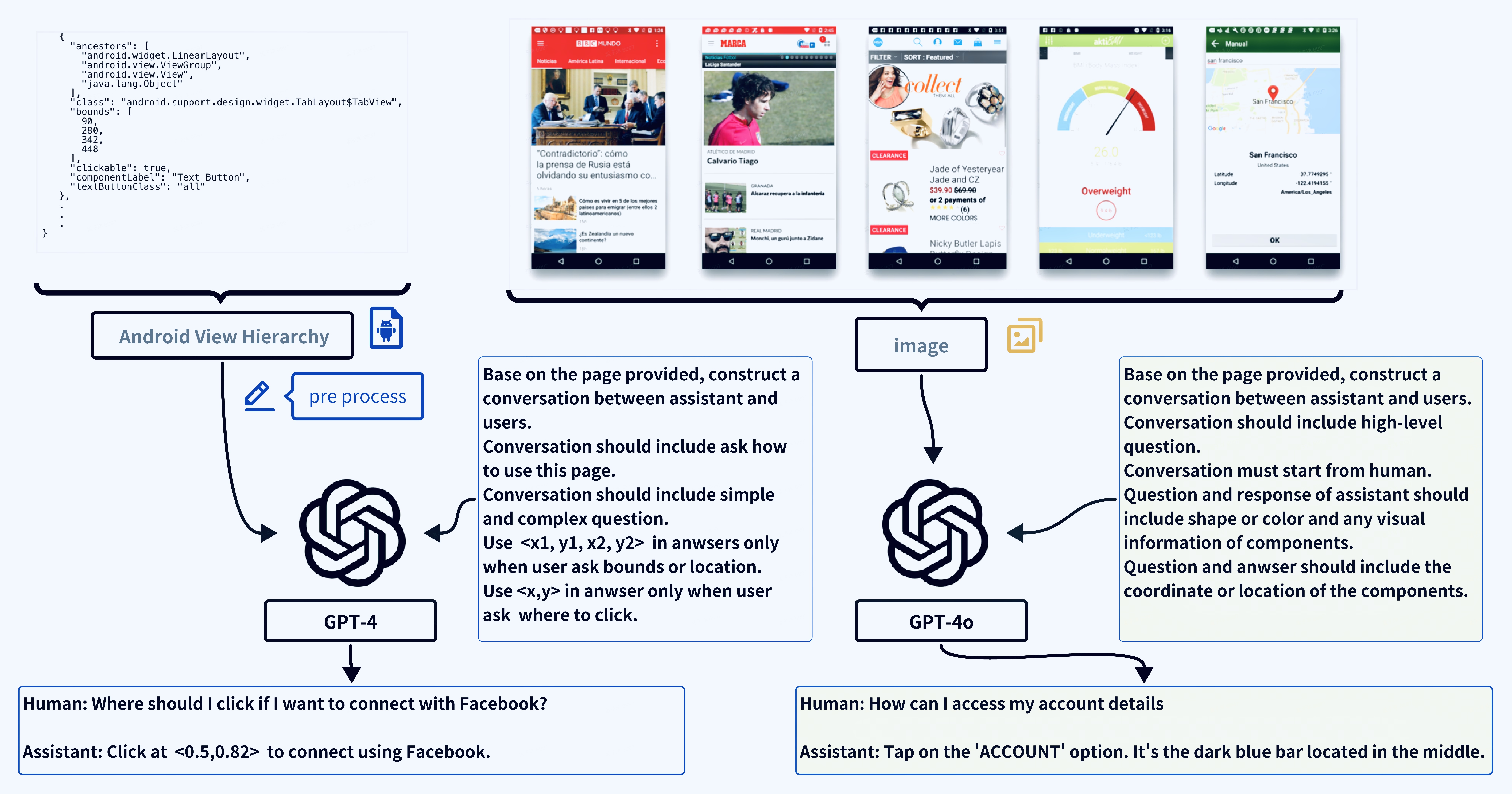} 
    \caption{Data generation Method}
    \label{fig:datageneration}
\end{figure*}

\section{Problem in GUI Comprehension}

The primary cause of hallucinations in Vision-Language Models (VLMs) can be attributed to their reliance on response patterns learned from traditional Large Language Models (LLMs). In this section, we investigate the occurrence of hallucinations in contemporary GUI models and propose a method to reduce these inaccuracies.

\subsection{Hallucination in GUI Comprehension}
\label{sec:reference}
LLMs are typically trained on massive corpora of pure text data, where they learn to generate responses based on textual context and patterns. When a visual module is introduced, these learned textual response patterns can adversely influence the model's behavior during multimodal tasks. 

Indeed, while Ferrent-UI and CogAgent are adept at understanding specific elements in graphical interfaces, they primarily emphasize textual and positional information when constructing their training datasets. This approach results in limited recognition capabilities for icons and other crucial visual details. Such a bias can lead to misinterpretations or incomplete understanding of GUIs that are visually complex in real-world applications, as reliance on text and layout alone may not adequately capture the full semantics and functionalities of the graphical interfaces. This models can face several types of illusions as follow:          


\textbf{Over-reliance on textual content}: LVLMs like Ferrent-UI and CogAgent, heavily trained on text-intensive datasets and focused on text interpretation tasks, tend to overly prioritize textual data when analyzing GUIs. As a result, they might overlook integral visual cues like layout and visual styles, leading to impractical or irrelevant action suggestions within the GUI environment, as shown in Table~\ref{tab:hallucination}. This text bias affects their capability to deliver effective and holistic interpretations, which is crucial for accurate and user-friendly interface interactions.

\textbf{Word-to-image coincidences}: This issue arises when models draw superficial connections between query words and text appearing on GUI elements, but these elements do not align with the intended functional requirements. As shown in Table~\ref{tab:case2}, if a query includes "start painting," and there are two buttons on the GUI with "start now" and "drawing" in their labels, a model might incorrectly choose a "start now" button when the appropriate action was to "drawing". This misalignment leads to the selection of incorrect actions based on text matches rather than the functional relevance of the elements within the GUI context.



\noindent
\subsection{Referent Method}\label{sec:reference}
To address the previously mentioned issue, we propose the \textit{referent method}. GUI design often uses layout, shape, and color to distinguish elements, providing critical visual details to guide user interactions. By explicitly incorporating coordinates, shapes, colors, and relative positions between elements in the construction of our dataset, we aim to enhance the model's focus on the image content when generating answers. Thereby reducing the chances of hallucination. During the design of our dataset, we ensured that most responses involving GUI elements include at least one of the following referents, aligning the element information with the image content:

\noindent
\textbf{ Shape}: The shape of the element, like a rectangular button with rounded corners.

\noindent
\textbf{ Color}: The color of the element, for example, a blue button with white text.

\noindent
\textbf{ Position}: The exact coordinates or bounds of the element, format as <x,y> and <x1,y1,x2,y2>.

\noindent
\textbf{ Relative Position}: The position of the element in relation to other elements, such as being below the text input field.

\section{GUI Comprehension Dataset}
To address the imbalance between image and text in GUI analysis, we propose a high-quality comprehension dataset tailored for LVLM training. This dataset includes detailed annotations of both textual and visual elements, ensuring a balanced and comprehensive training environment.

\subsection{Existing General Dataset}

Ferret-UI \cite{you2024ferretuigroundedmobileui} and CogAgent \cite{hong2023cogagentvisuallanguagemodel}, while capable of understanding certain elements in graphical interfaces, primarily focus on extracting and utilizing textual and positional information during the construction of their training datasets. This bias can lead to misunderstandings or inadequate interpretations when dealing with GUIs rich in visual information in actual applications, as relying solely on text and layout might not fully comprehend the semantics and functionalities of the graphical interfaces.

Additionally, since the datasets relied upon by these models are not open to the public, it poses additional challenges for external researchers or developers, as they might struggle to obtain data of similar quality and scale for effective model fine-tuning or further research.

\subsection{Data collection}

Given that both Ferrent-UI and CogAgent are not open source, the availability of the Rico dataset \cite{10.1145/3126594.3126651} represents a valuable resource for the research community involved in mobile app design and development.

The Rico \cite{10.1145/3126594.3126651} dataset is a huge GUI dataset created to support research in mobile app design and development, including areas such as GUI design, interaction. It consist of 66k unique GUI screens and 3M elements from 27 categories, over 9.3k applications. Each GUI comprise a screenshot and an augmented Android view hierarchy that capture all of the elements comprising a GUI, their properties, and relationships between them. However, the rich data provided by Rico cannot be directly utilized to train LVLMs as the text data in Rico do not align with human perception in the same manner as VQA datasets. This necessitates the transformation of this data into a format compatible with VQA dataset.

\begin{table*}
\centering
\begin{tabular}{p{2cm} c p{1.5cm} c p{1.5cm} c p{2cm} c p{1.5cm} c p{1.5cm} c}
\hline
\fontsize{10}{12}\selectfont{\textbf{Inst dataset}} & \fontsize{10}{12}\selectfont{\textbf{Model}} & \fontsize{10}{12}\selectfont{\textbf{Number}}& \fontsize{10}{12}\selectfont{\textbf{Conv dataset}} & \fontsize{10}{12}\selectfont{\textbf{Model}} & \fontsize{10}{12}\selectfont{\textbf{Number}}\\
\hline
\fontsize{8}{10}\selectfont{description\_inst}      &  \fontsize{8}{10}\selectfont{GPT-4}   & \fontsize{8}{10}\selectfont{3k} & \fontsize{8}{10}\selectfont{Conv\_simple}      &  \fontsize{8}{10}\selectfont{GPT-4}   & \fontsize{8}{10}\selectfont{5.4k}\\
\fontsize{8}{10}\selectfont{bound\_inst}          &  \fontsize{8}{10}\selectfont{GPT-4}   & \fontsize{8}{10}\selectfont{4.3K} & \fontsize{8}{10}\selectfont{Conv\_complex}          &  \fontsize{8}{10}\selectfont{GPT-4}   & \fontsize{8}{10}\selectfont{11K} \\
\fontsize{8}{10}\selectfont{function\_inst}        & \fontsize{8}{10}\selectfont{GPT-4}    & \fontsize{8}{10}\selectfont{2K} & \fontsize{8}{10}\selectfont{Conv\_4o\_long}             & \fontsize{8}{10}\selectfont{GPT-4o}   & \fontsize{8}{10}\selectfont{10K} \\
\fontsize{8}{10}\selectfont{testing\_inst}     & \fontsize{8}{10}\selectfont{GPT-4}    & \fontsize{8}{10}\selectfont{5K} & \fontsize{8}{10}\selectfont{Conv\_4o\_short}     & \fontsize{8}{10}\selectfont{GPT-4o}    & \fontsize{8}{10}\selectfont{10K}\\
\fontsize{8}{10}\selectfont{function\_inst\_4o}    & \fontsize{8}{10}\selectfont{GPT-4o}   & \fontsize{8}{10}\selectfont{8K} & \fontsize{8}{10}\selectfont{conv\_4o\_miss} & \fontsize{8}{10}\selectfont{GPT-4o} & \fontsize{8}{10}\selectfont{5k} \\
\hline
\fontsize{8}{10}\selectfont{Total}                   &          &    &                   &          &  \fontsize{8}{10}\selectfont{63.8K}  \\
\hline
\end{tabular}
\caption{Dataset composition}
\label{dataset-compose}
\end{table*}

\subsection{Task Design} \label{sec:ftask}
The dataset includes the following sub-tasks, each complemented by human annotations to elevate the quality and applicability of the training data. Additionally, we have structured the dataset into two distinct categories based on these tasks: the instruction dataset and the conversation dataset. These cater to scenarios of directive compliance and multi-turn dialogue, respectively.

\begin{itemize}
    \item \textit{Description}: This task involves providing a basic description of the GUI's layout, identifying the function and placement of components. This foundational task facilitates deeper analytical tasks, as all GUI-related tasks rely on a clear understanding of the overall layout.
    \item \textit{Bounds \& Location}: This task adds complexity by requiring the model to incorporate precise descriptions of element bounds and coordination. By introducing this, our approach ensures that the model focuses on the accurate positional information within the images.
    \item \textit{Function}: This task requires the model to understand the individual functions of each GUI element, considering not only the overall function but also the relative positions of elements to each other. Identical buttons might serve different purposes depending on their specific context within a GUI. while relative positioning is significant because two elements placed in a specific arrangement might together convey a particular function or information. This enables the model to predict the design intent and functionality of the entire interface, resulting in more contextually aware and precise responses.
\end{itemize}

\subsection{Generation Method}
However, the rich data provided by Rico cannot be directly utilized to train LVLMs as the text data in Rico do not align with human perception in the same manner as VQA datasets. This necessitates the transformation of this data into a format compatible with VQA dataset. The Android view hierarchy in the Rico dataset provides detailed descriptions of GUI elements, allowing us to obtain precise positional data and interaction details. Therefore, we use GPT-4 (text-only) to extract this textual information from the Android view hierarchy and GPT-4o to extract visual information from images as shown in Figure~\ref{fig:datageneration}. Our approaches are as follows:

\noindent
 \textit{Data Pre-processing}: We remove elements in the Android View Hierarchy that have the `visible-to-user' attribute set to `false', and normalize the `bounds' values to a range of 0 to 1. Additionally, for all elements with the `clickable' attribute set to `true', we add a `click\_coordinate' attribute, calculated as the midpoint of the range indicated by the `bounds'.

 \noindent
\textit{GPT-4 (Text-Only) Generated Tasks}:  Inspired by the method of LLaVA \cite{liu2024visual}, this category consists of dataset generated by GPT-4 (text-only), based on pre-processed Android View Hierarchy data. Leveraging the detailed information available in the Android View Hierarchy, this approach generates a variety of QA pairs that accurately reflect the textual and positional data and interactions.

\noindent
\textit{GPT-4o (Image-Based) Generated Tasks}: Inspired by ShareGPT4 \cite{chen2023sharegpt4v}, this category comprises dataset generated by GPT-4o (image-based). Given that vision models often struggle with capturing precise positional information of elements, this method focus on relative positions, shapes, and colors of the visual elements.

Our GUI comprehension dataset consists of 63.8k images, covering a diverse range of apps and tasks. It includes 22.3k instruction-following data pairs and 41.4k conversation data pairs. Of the instruction data, 35.8\% is generated by GPT-4o, and 60.2\% of the conversation data is produced by GPT-4o. Table~\ref{dataset-compose} illustrates the composition of our dataset.

\section{Tuning Script}\label{sec:finetune}
Existing fine-tuning methods in LVLM treat question tokens and image tokens equally, which neglects the need for LVLM to prioritize image information. As a result, models may understand tasks well but generate responses based solely on questions. This results in hallucinations in GUI comprehension (see Table~\ref{tab:hallucination}). To address this, we propose a two-stage training method. The first stage aligns responses with image content. The second stage aligns responses with human intent. We also adopt the chain-of-thought method to enhance the model's reasoning capability. Our approaches are as follows:

\textbf{Foundation Stage}: 

The Foundation Task dataset is an instruction-following dataset that fix the command and response formats. This stage incorporates direct visual information and employs a fixed format for questions and responses to train the model to correlate with image content. The fixed format is intended to "freeze" the variations in questioning and responding styles, focusing the model’s learning process on understanding and interpreting visual attributes effectively. This controlled environment ensures that the model develops a robust capability to recognize and interpret visual data independently from textual content.

\textbf{Advanced Comprehensive Stage}:  

In this stage, we introduce multi-turn dialogues and complex questions from the advanced comprehension dataset. We utilize the Referent Method (detailed in 
sec~\ref{sec:reference}) to directly incorporate intuitive visual information into responses. This method is designed not only to enhance the referencing of visual details within responses but also to align the model’s attention distribution more towards image tokens.It ensures that the model does not merely rely on learned text patterns but actively engages with and interprets the visual context, enabling a deeper understanding and more accurate generation based on the actual GUI layout and visual cues.

\begin{figure}
    \centering
    \includegraphics[width=0.5\textwidth]{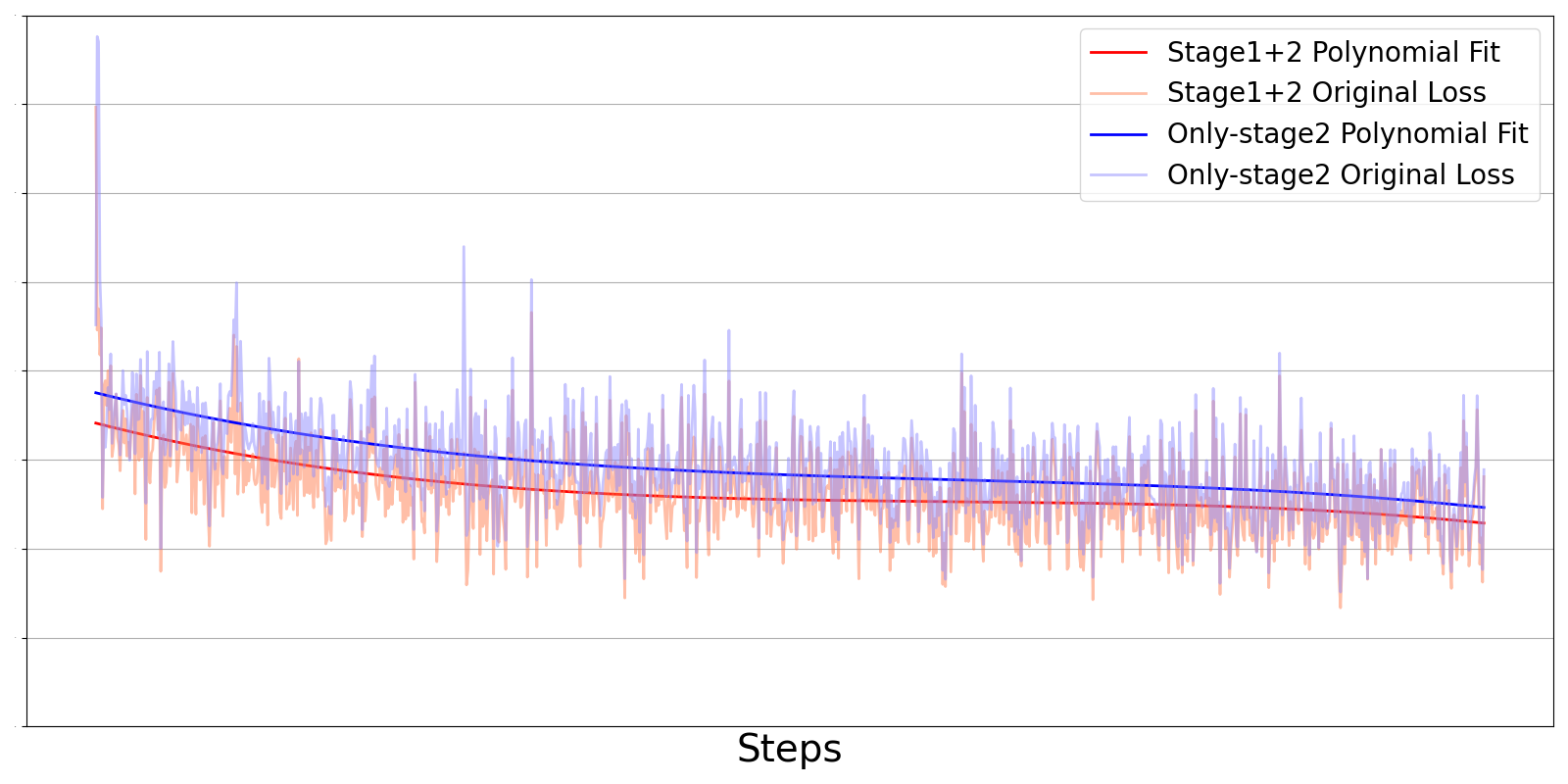} 
    \caption{Loss convergence of model trained with foundation task and without during advanced task training.}
    \label{fig:covergence}
\end{figure}

\begin{figure}
    \centering
    \includegraphics[width=0.5\textwidth]{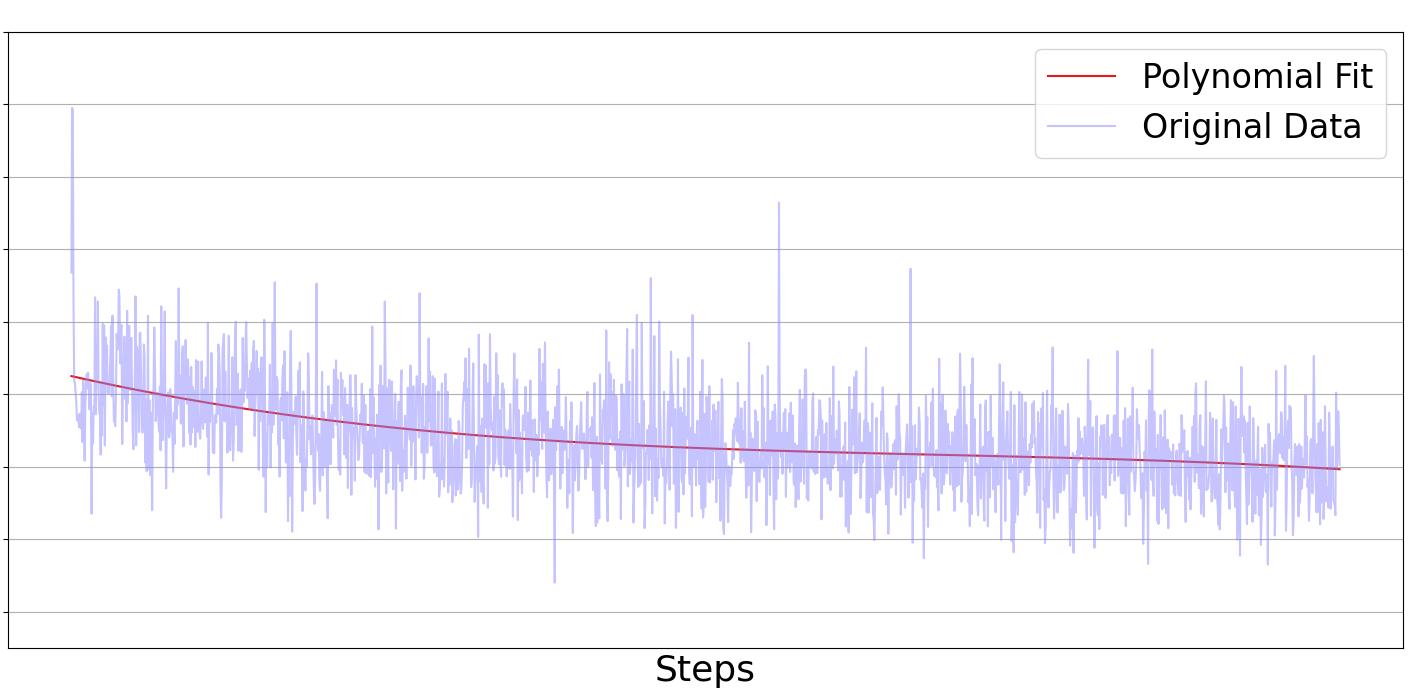} 
    \caption{Loss convergence of model trained with mixed foundation task and advanced task data.}
    \label{fig:covergence-all}
\end{figure}


\textbf{Task Progression and Reinforcement}: 

We organize tasks in a sequence from simple to complex, ensuring that the model progressively builds the necessary foundational skills before training on more advanced tasks.
In complex tasks, we start with some related simple tasks to reinforce foundational knowledge, followed by the real response which demands deeper analysis and synthesis of the information gathered from these foundational tasks.

\begin{table}[h]
    \centering
    \begin{tabular}{p{1.5cm} p{1cm} p{1cm} p{1cm} p{1cm}}
    \hline
         \fontsize{10}{10}\selectfont{\textbf{Model}} & \fontsize{10}{10}\selectfont{\textbf{First}} & \fontsize{10}{10}\selectfont{\textbf{Second}} &  \fontsize{10}{10}\selectfont{\textbf{Third}} & \fontsize{10}{10}\selectfont{\textbf{Average}}\\
    \hline
         \fontsize{8}{10}\selectfont{{VGA-no-referent}}& \fontsize{8}{10}\selectfont{50.68} & \fontsize{8}{10}\selectfont{48.18} &\fontsize{8}{10}\selectfont{46.14} & \fontsize{8}{10}\selectfont{48.33}\\
    \hline
         \fontsize{8}{10}\selectfont{{VGA-mix-stage}}& \fontsize{8}{10}\selectfont{62.05} & \fontsize{8}{10}\selectfont{62.96} &\fontsize{8}{10}\selectfont{62.27} & \fontsize{8}{10}\selectfont{62.42}\\
    \hline
        \fontsize{8}{10}\selectfont{{VGA-only-stage2}}& \fontsize{8}{10}\selectfont{67.27} & \fontsize{8}{10}\selectfont{65.00} &\fontsize{8}{10}\selectfont{67.50} & \fontsize{8}{10}\selectfont{66.59}\\
    \hline
         \fontsize{8}{10}\selectfont{VGA-7b-v1}& \fontsize{8}{10}\selectfont{\textbf{90.68}} & \fontsize{8}{10}\selectfont{\textbf{90.68}} &\fontsize{8}{10}\selectfont{\textbf{91.17}} & \fontsize{8}{10}\selectfont{\textbf{90.83}}\\
    \hline
    \end{tabular}
    \caption{We conducted ablation experiments on our fine-tuning methods, and the results prove that our fine-tuning methods are effective.}
    \label{tab:duibi}
\end{table}




\begin{table}
    \centering
    \begin{tabular}{p{1.5cm} p{1cm} p{1cm} p{1cm} p{1cm}}
    \hline
         \fontsize{10}{10}\selectfont{\textbf{Model}} & \fontsize{10}{10}\selectfont{\textbf{First}} & \fontsize{10}{10}\selectfont{\textbf{Second}} &  \fontsize{10}{10}\selectfont{\textbf{Third}} & \fontsize{10}{10}\selectfont{\textbf{Average}}\\
    \hline
         \fontsize{8}{10}\selectfont{GPT-4o}& \fontsize{8}{10}\selectfont{80.68} & \fontsize{8}{10}\selectfont{80.45}&\fontsize{8}{10}\selectfont{81.14}&\fontsize{8}{10}\selectfont{80.75}\\
         \fontsize{8}{10}\selectfont{GPT-4V}& \fontsize{8}{10}\selectfont{81.82} & \fontsize{8}{10}\selectfont{81.14} & \fontsize{8}{10}\selectfont{82.50}&\fontsize{8}{10}\selectfont{81.82}\\
         \fontsize{8}{10}\selectfont{MiniCPM} & \fontsize{8}{10}\selectfont{63.86} & \fontsize{8}{10}\selectfont{64.09} &\fontsize{8}{10}\selectfont{64.32}& \fontsize{8}{10}\selectfont{64.09}\\
         \fontsize{8}{10}\selectfont{CogAgent} & \fontsize{8}{10}\selectfont{69.77} & \fontsize{8}{10}\selectfont{69.77} &\fontsize{8}{10}\selectfont{69.09}& \fontsize{8}{10}\selectfont{69.55}\\
         \fontsize{8}{10}\selectfont{llava-next} & \fontsize{8}{10}\selectfont{53.41} & \fontsize{8}{10}\selectfont{53.18}&\fontsize{8}{10}\selectfont{53.86}& \fontsize{8}{10}\selectfont{53.48}\\
         \fontsize{8}{10}\selectfont{idefics2-8b} & \fontsize{8}{10}\selectfont{39.77} & \fontsize{8}{10}\selectfont{43.86} &\fontsize{8}{10}\selectfont{41.81}& \fontsize{8}{10}\selectfont{41.82}\\
    \hline
         \fontsize{8}{10}\selectfont{{VGA-7b-v1*}}& \fontsize{8}{10}\selectfont{59.00} & \fontsize{8}{10}\selectfont{59.75} &\fontsize{8}{10}\selectfont{57.75} & \fontsize{8}{10}\selectfont{58.5}\\
         \fontsize{8}{10}\selectfont{VGA-7b-v1}& \fontsize{8}{10}\selectfont{\textbf{90.68}} & \fontsize{8}{10}\selectfont{\textbf{90.68}} &\fontsize{8}{10}\selectfont{\textbf{91.17}} & \fontsize{8}{10}\selectfont{\textbf{90.83}}\\
    \hline
    \end{tabular}
    \caption{Scores on GUI Bench.}
    \label{tab:bench}
\end{table}

\subsection{Experiments Setup}
We use our training method and dataset to fine-tune llava-v1.6-mistral-7b \cite{liu2024llavanext}. Table~\ref{training-parament} shows the hyperparameters we use during training. In Appendix~\ref{Hypeanalysis}, we show the loss convergence behavior under various learning rates and batch sizes during training. All our experiments are carried out using one A100(80GB) machine.



\begin{table*}
    \centering
    \begin{tabular}{p{2.5cm}p{5cm}p{5cm}}
        \hline
        \textbf{\fontsize{12}{14}\selectfont Case in bench} & & \\
        \hline
        & \includegraphics[width=0.15\textwidth]{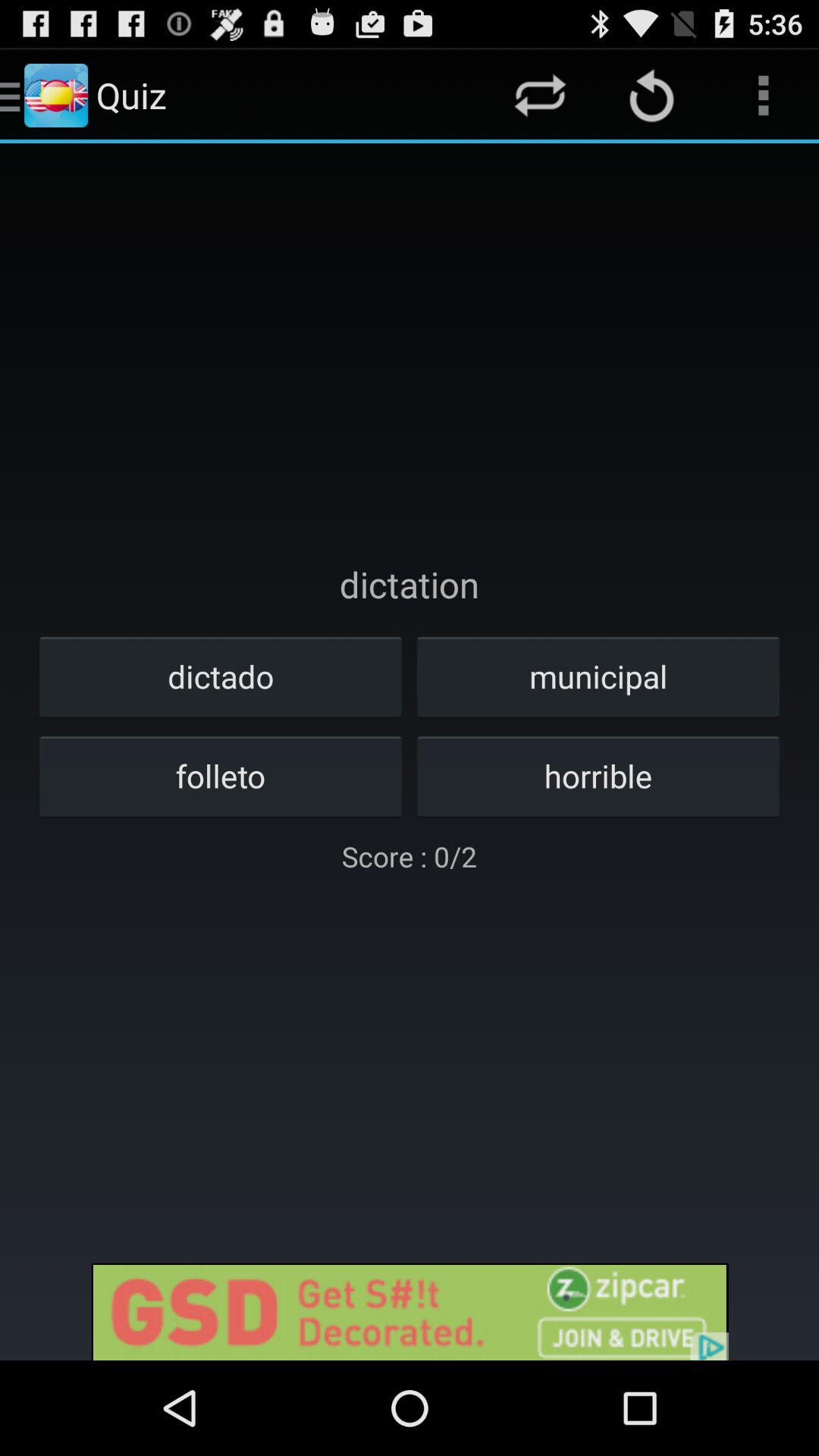} & \includegraphics[width=0.15\textwidth]{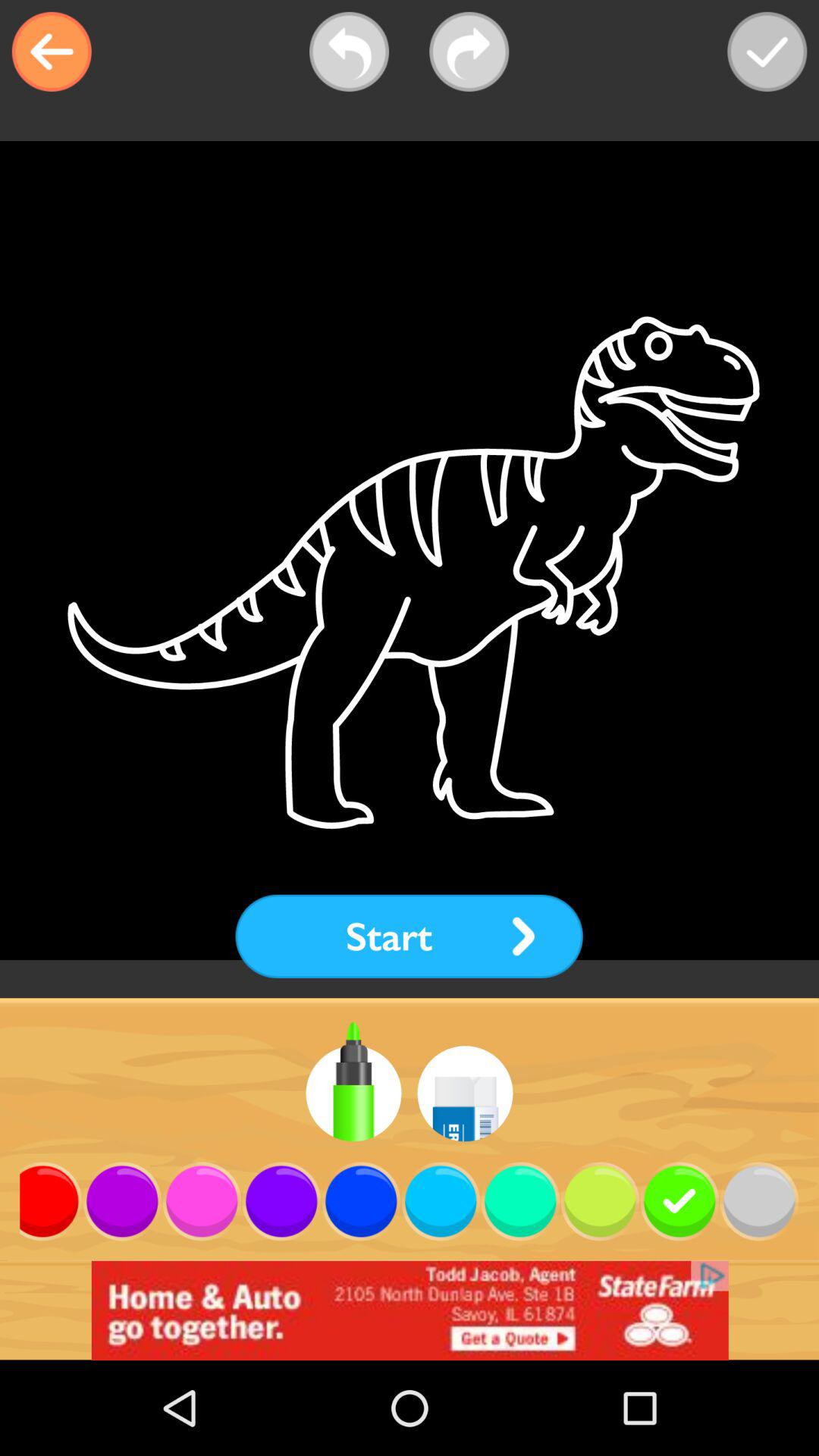}\\
        \fontsize{8}{10}\selectfont{User} & \fontsize{8}{10}\selectfont{\textcolor{blue}{How can I score?}} & \fontsize{8}{10}\selectfont{\textcolor{blue}{How can I draw a picture?}} \\
        \fontsize{8}{10}\selectfont{VGA-7b-v1} & \fontsize{8}{10}\selectfont{You can score the \textcolor{teal}{`dicatado' button}, which is located in the middle left of the page.} & \fontsize{8}{10}\selectfont{You can click the \textcolor{teal}{blue `Start' button} in the center of the page to begin drawing a picture.} \\
        \fontsize{8}{10}\selectfont{VGA-7b-v1*} & \fontsize{8}{10}\selectfont{You should \textcolor{teal}{click the `dictado' button} located in the top left of the four options.} & \fontsize{8}{10}\selectfont{You can start drawing by clicking the \textcolor{teal}{blue 'Start' button}} \\
        \hline
        \fontsize{8}{10}\selectfont{llava-mistral-v1.6-7b} & \fontsize{8}{10}\selectfont{The click to score is the one that says \textcolor{red}{"horrible"}} & \fontsize{8}{10}\selectfont{To draw a picture, you can use a \textcolor{red}{drawing app} on your phone or tablet.} \\
        \hline
        \fontsize{8}{10}\selectfont{Idefics2-8B} & \fontsize{8}{10}\selectfont{To get the score, you should click on the \textcolor{red}{"Score"} button.} & \fontsize{8}{10}\selectfont{To draw a picture, simply \textcolor{red}{tap on the colors you want to use and then touch the screen where you would like to place the color}...} \\
        \hline
        \fontsize{8}{10}\selectfont{MiniCPM-V-2\_5} & \fontsize{8}{10}\selectfont{To score, you click on the word .\textcolor{red}{'horrible'}} & \fontsize{8}{10}\selectfont{To draw a picture, you can use the \textcolor{red}{drawing tools provided in the app interface}.} \\
        \hline
    \end{tabular}
    \caption{Case in the GUI-bench, note that VGA show the best comprehension of the image}
    \label{tab:case}
    
\end{table*}

\section{Experiment}
\subsection{Baselines \& Evaluation Metric}

\textbf{GUI comprehension Bench}: To evaluate our model, we follow previous works \cite{masry2023unichart, liu2023improved}. Due to the lack of GUI comprehension bench, we sample 22 images from the Rico dataset (excluding training data). And based on these images, we collect 44 user questions which require truly understanding of GUI to response correctly. Inspired by the evaluation method of LLaVA-bench (in-the-wild) \cite{liu2024visual}, we use ChatGPT to evaluate our model. We compare our model with five recent best performance LVLMs: the top non-open-source LVLMs GPT-4V and GPT-4o, the two best LVLMs based on Mistral-7b-instruction-v2 (llava-v1.6-mistral-7b\cite{liu2024llavanext} and idefics2-8b\cite{laurençon2024matters}), and MiniCPM-llama3-V-2.5\cite{hu2024minicpm}, the leading open-source model on the LVLM leaderboard.

\textbf{Fine-tuning Method Evaluation}: To evaluate our two-stage fine-tuning approach (Foundation and Advanced Comprehension method), we track loss convergence over the training period. We compare loss convergence during the advanced task training between models pre-trained with the foundation task and which without it. We also compare previous models' loss convergence with which trained on mixed foundation task and advanced task data. Their learning rates during training are all set to 2e-5, batch sizes are 16.

\subsection{Main Result}
\textbf{GUI Comprehension}: As shown in Table~\ref{tab:bench}, VGA-7b-v1 has shown promising results, achieving the best performance across three separate GPT evaluations. Our model attains a score of 90.83, which is relatively 41\% better than the base model llava-v1.6-mistral-7b, 54\% better than idefics2-8b and 29\% better than MiniCPM-llama3-V-2.5. Furthermore, VGA-7b-v1 also outperforms GPT-4o and GPT-4V. Overall, these results establish VGA as the SOTA model for GUI comprehension.

\textbf{Resource Efficiency}: We also evaluate our model's performance with low-resolution input. As shown in Table~\ref{tab:bench}, VGA-7b-v1* (336x336) still outperforms llava-v1.6-mistral-7b and idefics-8b, even though the number of image tokens for VGA-7b-v1* is only one-fourth of those in llava-v1.6-mistral-7b and idefics-8b. This significantly improves the inference speed, computational cost and memory usage.
Our research indicates that reducing pixel count primarily affects detail recognition accuracy, thereby impacting response accuracy. Nevertheless, the model retains a strong understanding of GUI images, as demonstrated in Table~\ref{tab:case}.

\textbf{Hallucination Analysis}: Compared with existing models, our model demonstrates a precise understanding of GUI. As shown in Table~\ref{tab:hallucination} and Table~\ref{tab:case2}, our model responds to questions based on the input image content and accurately describe GUI components using details such as relative position, color, and shape. By focusing more on image content, our model certainly reduces the likelihood of generating hallucinated responses.



\textbf{FAC method analysis}: As shown in Figure~\ref{fig:covergence}, the orange and purple curves represent the original loss values for models with and without the foundation task stage, respectively. The red and blue curves, fitted with a degree-3 polynomial, illustrate the loss trends. Models trained with the foundation task exhibit more stable convergence and lower loss values. Figure~\ref{fig:covergence-all} shows that mixing datasets during training leads to unstable convergence. In Table~\ref{tab:duibi}, we present the results of ablation experiments on the GUI comprehension bench using the FAC fine-tuning method and those not using the FAC method. The experiment proves that the models using the FAC method have stronger GUI understanding capabilities, consistent with the comparison of the loss curves. In Appendix~\ref{Appendix:Stageanalysis}, we discuses the influence of the foundation stage on model convergence. In Appendix~\ref{appendix:ablation}, we conducted ablation studies to evaluate the individual contributions of the two stages to overall model performance. In Appendix~\ref{appendix:attention}, we analyzed the contribution of image tokens when generating answers to demonstrate how FAC enhances the model's attention to image content.


\section{Conclusion}
We introduce VGA, a fine-tuned model for GUI comprehension, leveraging a custom dataset and novel fine-tuning method. The dataset, constructed using a \textit{Reference Method}, focuses on image-centric responses. The training can be divided into two stage: foundational skill alignment and human intent alignment. The foundation stage emphasizes learning the relationship between images and responses, while the advanced stage focuses on intent following. Ablation studies confirm the efficacy of our training method. We believe that our training data and method will serve as valuable resources for future research.

\section*{Limitations}
Despite the promising results, our work has some limitations that need to be addressed in future research.
Firstly, our dataset contains some noisy data, which may affect the overall performance of the models. We plan to clean this data in future iterations to improve the quality and accuracy of the dataset.
Secondly, the accuracy of responses is somewhat limited by the capability of the base model we used. Although our constructed dataset and proposed methodologies have demonstrated their effectiveness, there are still instances where the answers may be inaccurate. Future work will involve experimenting with more advanced base models to further validate and enhance the effectiveness of our approach.
Due to the commercial licensing restrictions of the LLaMA model, we were unable to fine-tune based on the LLaMA model. At the time of our work, mistral-7b-instruction-v3 had not yet been released. Therefore, we utilized llava-v1.6-mistral-7b. In our research, it is one of the most advanced open-source and commercially viable models available.
Thirdly, our dataset is constructed based on the open-source RICO dataset, which contains GUI images with older design styles. In future dataset construction, we aim to integrate GUI interfaces from current popular apps to ensure our models are up-to-date with modern design trends and practices. By addressing these limitations, we aim to refine our approach and further improve the effectiveness and reliability of our models in comprehending and interacting with complex GUI interfaces.




\section*{Ethics Statement}
During the development process, we ensured compliance with the terms and conditions of the various models and datasets we utilized. The foundation model we employed, llava-v1.6-mistral-7b, is based on the mistral-7b-instruction-v2 and CLIP models. Both llava-v1.6-mistral-7b and mistral-7b-instruction-v2 are licensed under the Apache 2.0 License, which grants open-source and commercial-use permissions from both llava and mistral AI.
Our dataset was constructed based on the RICO dataset. The University of Illinois provides a permissive license for the RICO dataset under the condition that users comply with their terms, allowing for open-source and commercial use.
Due to the generative nature of our models, there is a potential risk that they may be misused to generate factually incorrect responses that could misinform the public. Additionally, we cannot guarantee that our models will not produce text containing hate speech or harmful content.

\section*{Acknowledgements}
The authors would like to thank the anonymous reviewers for their helpful comments. This research was supported by Bytedance, we would like to extend our gratitude to Bytedance for their assistance and support.



\appendix

\section{Appendices}

\subsection{Existing Method For Data Generation }
\label{knowledgedestill}
To effectively train an LVLM, an extensive amount of data with high-quality is required \cite{zhu2023minigpt4}. Due to the huge labor consumption to create a large high-quality dataset, many research initiatives have explored methods for automatic generation of datasets \cite{Gilardi_2023}\cite{vicuna2023}\cite{peng2023instruction}. 

Current mainstream method focus on distilling knowledge from advanced large language models and large visual-language models, which are pre-trained on vast amounts of data, often exceeding terabyte (TB) scales, to capture a wide variety of patterns in language and visuals \cite{xu2023wizardlm}\cite{vicuna2023}\cite{10.1145/3133920}\cite{chen2024allava}. Therefore, they are often used as auxiliary tools to generate high-quality data. For instance, 
ShareGPT4v \cite{chen2023sharegpt4v} distill the knowledge from GPT-4V to generate data for training, ensuring the image and text pairs are of high-quality and diversity.  LLaVA \cite{liu2023improved} turns pictures into text descriptions and then uses GPT-4 \cite{openai2024gpt4} to generate questions and answers based on text descriptions. Both of them get a state-of-the-art performance.

\subsection{Hyperparameter Analysis}
\label{Hypeanalysis}
Our fine-tuning process is based on the llava-v1.6-mistral-7b model. In this section, we examine the influence of different learning rates and batch sizes on model convergence across two stages. 

\begin{figure*}[h]  
    \centering
    \begin{minipage}{0.48\textwidth}
        \centering
        \includegraphics[width=\linewidth]{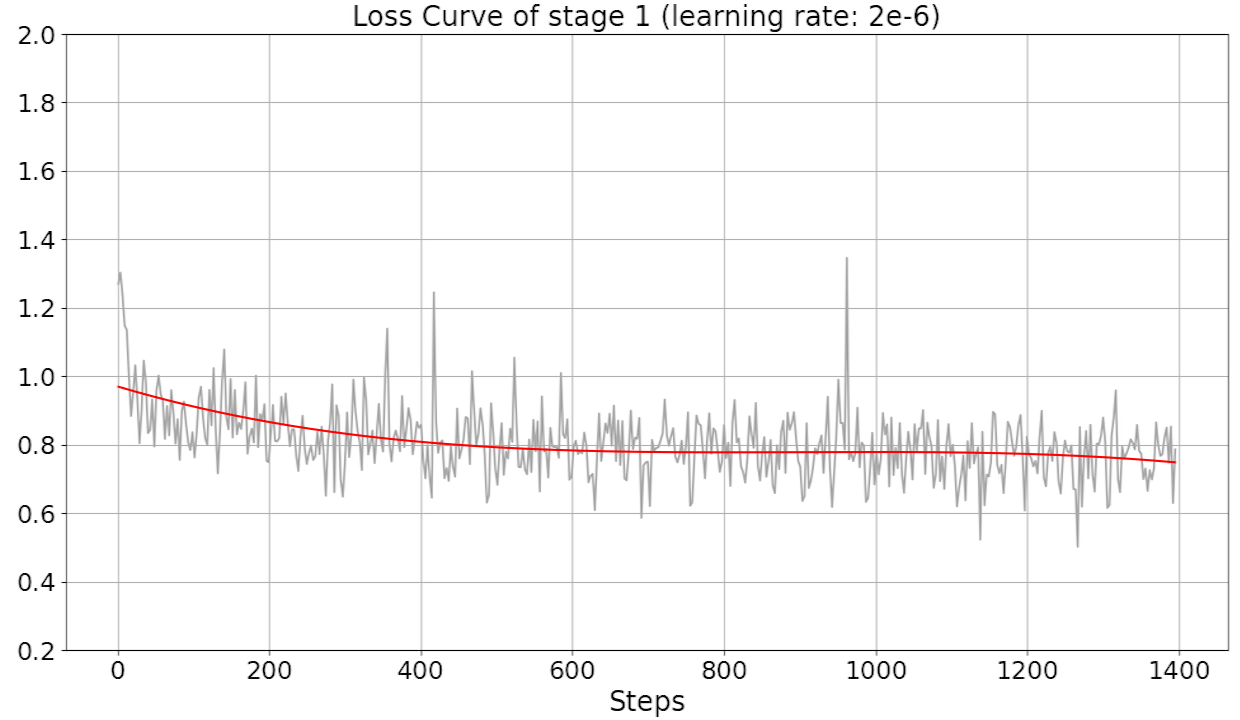} 
        \caption{}
        \label{stage1-5}
    \end{minipage}
    \begin{minipage}{0.48\textwidth}
        \centering
        \includegraphics[width=\linewidth]{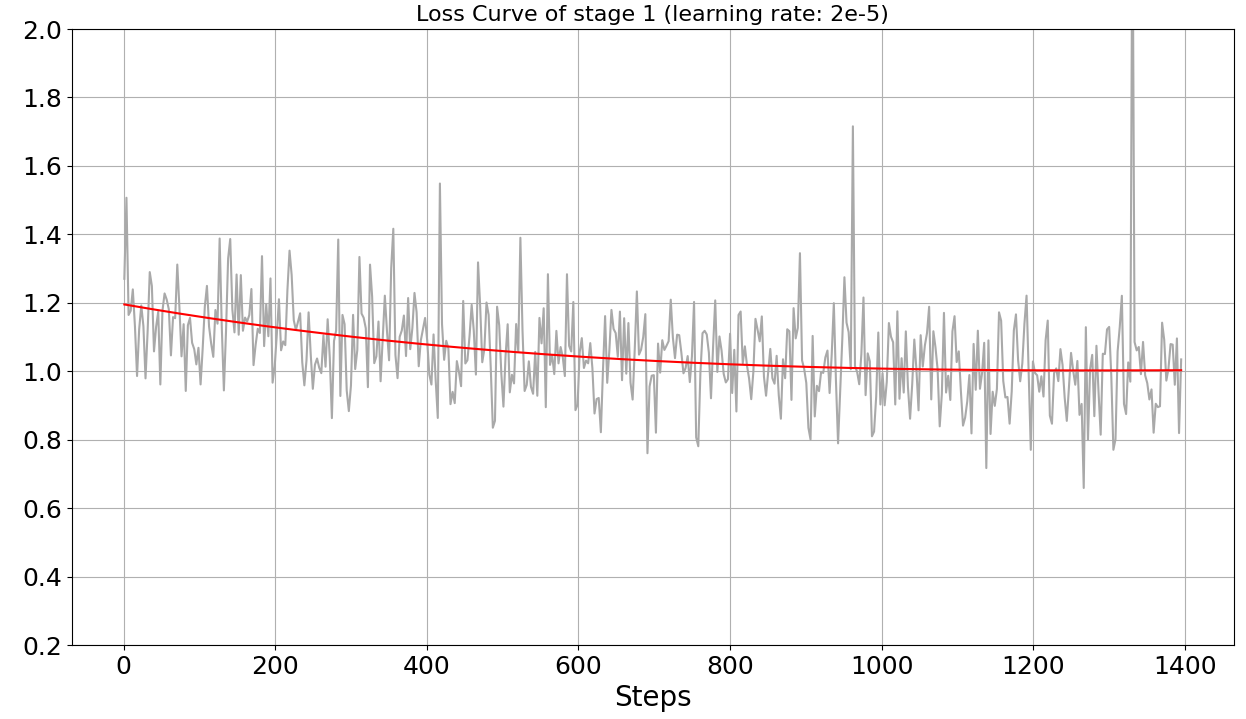} 
        \caption{}
        \label{stage1-6}
    \end{minipage}
    \begin{minipage}{0.48\textwidth}
        \centering
        \includegraphics[width=\linewidth]{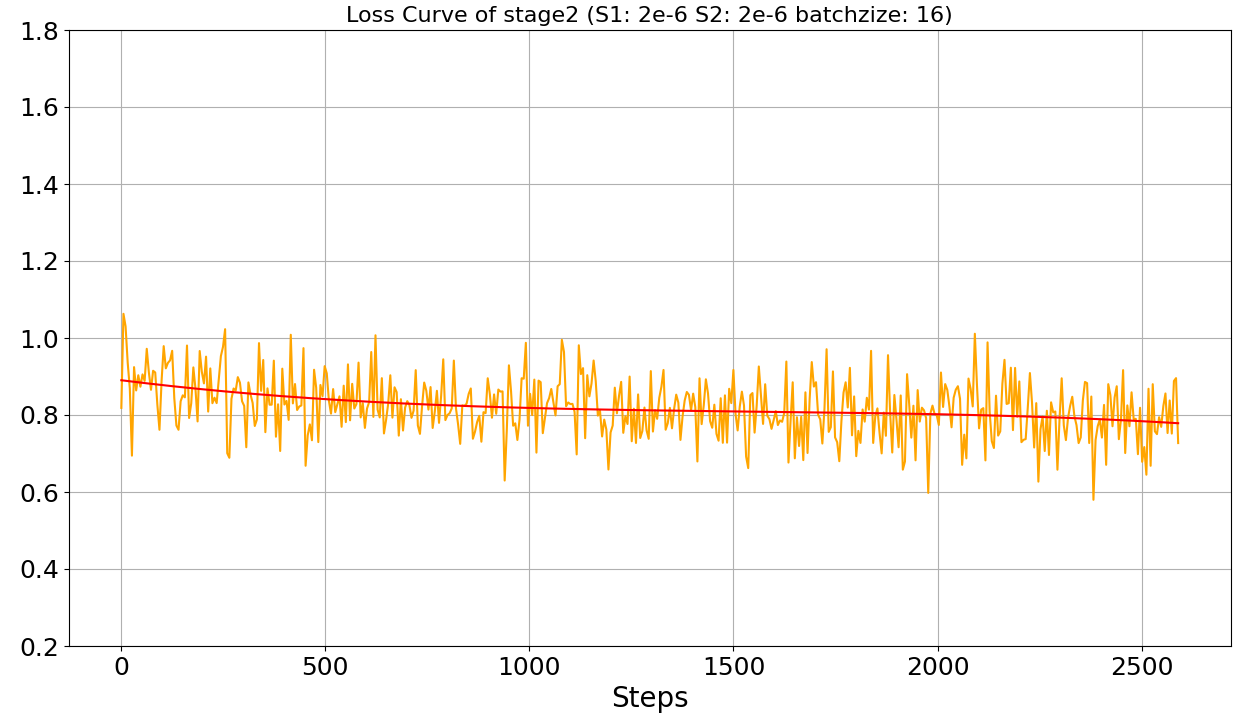} 
        \caption{}
        \label{v4}
    \end{minipage}
    \begin{minipage}{0.48\textwidth}
        \centering
        \includegraphics[width=\linewidth]{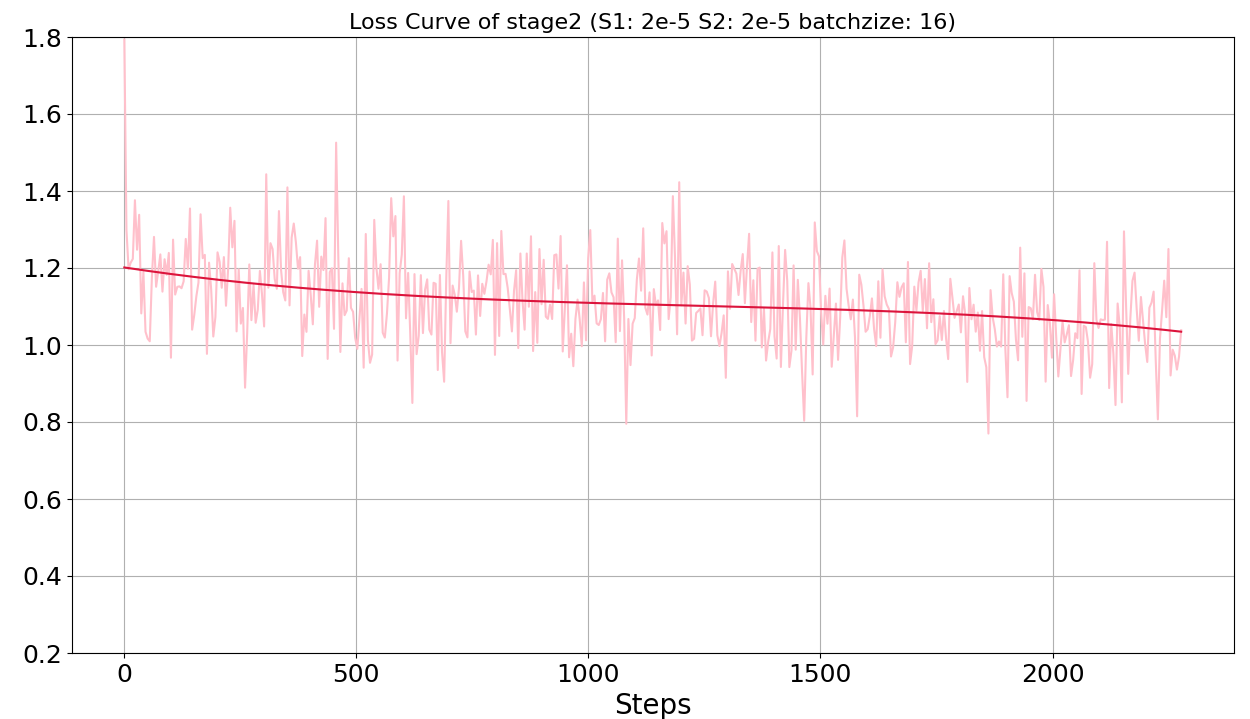} 
        \caption{}
        \label{v2}
    \end{minipage}
    \begin{minipage}{0.48\textwidth}
        \includegraphics[width = \textwidth]{UIllava-mistral-insconv-v6-loss.png}
        \caption{VGA-7b-v1}
        \label{v6}
    \end{minipage}
    \begin{minipage}{0.48\textwidth}
        \includegraphics[width = \textwidth]{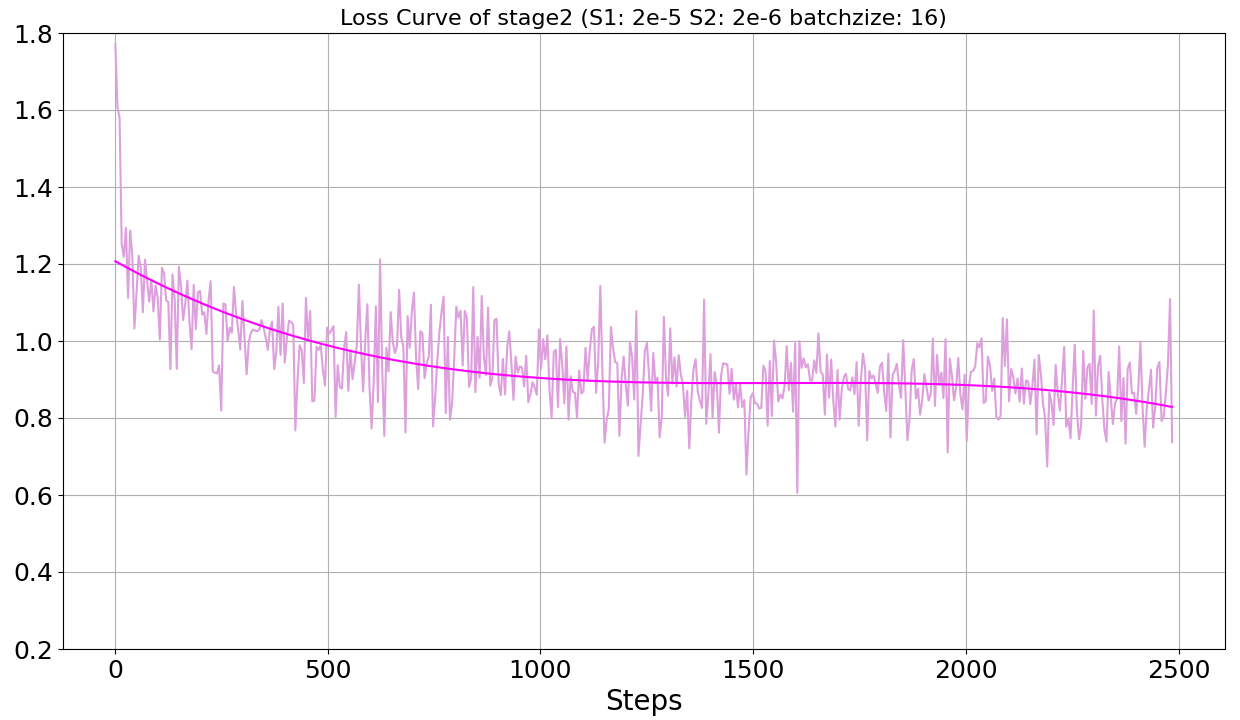}
        \caption{}
        \label{v5}
        \label{sec:appendex}
    \end{minipage}
\caption{Train hyperparameter analysis}
\label{lossanalysis}
\end{figure*}

Figures~\ref{stage1-5} and Figure~\ref{stage1-6} display the loss convergence during foundation task training at various learning rates, both of the batch sizes are 16. It is evident that a learning rate of 2e-6 achieves lower loss values and more stable learning, with significantly fewer fluctuations.

Figures~\ref{v4} and Figure~\ref{v5} show the loss convergence during advanced task training of models pre-trained with different learning rates on foundation task. The results indicate that model pre-trained with a learning rate of 2e-6 on foundation task results in markedly lower loss and reduced fluctuations during advanced task training.

Comparing Figure~\ref{v6} and Figure~\ref{v4}, which both pre-trained with learning rate of 2e-6 on foundation task, and with the different batch size (32 and 16). This comparison shows that increasing the batch size significantly improves convergence stability and reduces fluctuations during advanced task, avoids over-fit in intent following. 

In our experiments, the model trained under the conditions depicted in Figure~\ref{v6} surpasses others. This model achieves higher accuracy in GUI element recognition compared to the which pre-trained on foundation task with a learning rate of 2e-5. and show more accuracy in human intent following compared with model trained with smaller batch size during advanced task training. This further underscores the importance of foundation task training and highlights the distinct focuses of the two stages. Using a larger batch size in advanced helps the model capture the nuances of user intent, thereby preventing over-fitting. Using a small learning rate in foundation task training helps the model more accurately map images to their corresponding responses.

\subsection{Influence of Foundation Task}
\label{Appendix:Stageanalysis}
As shown in Figure~\ref{lossanalysis}, the performance during the foundation training phase significantly impacts the results of advanced task training. As shown in Figure~\ref{v4} and Figure~\ref{v5}, if the model achieves stable convergence during the foundation task, it will also exhibit stable convergence during the advanced task. Furthermore, the final convergence value in the advanced task is influenced by the convergence value achieved during the foundation task. Using models with lower convergence values from the foundation task leads to smaller convergence values during the advanced task task training.

\subsection{Ablation Study}
\label{appendix:ablation}
\begin{figure*}[h]  
    \centering
    \begin{minipage}{0.48\textwidth}
        \centering
        \includegraphics[width=\linewidth]{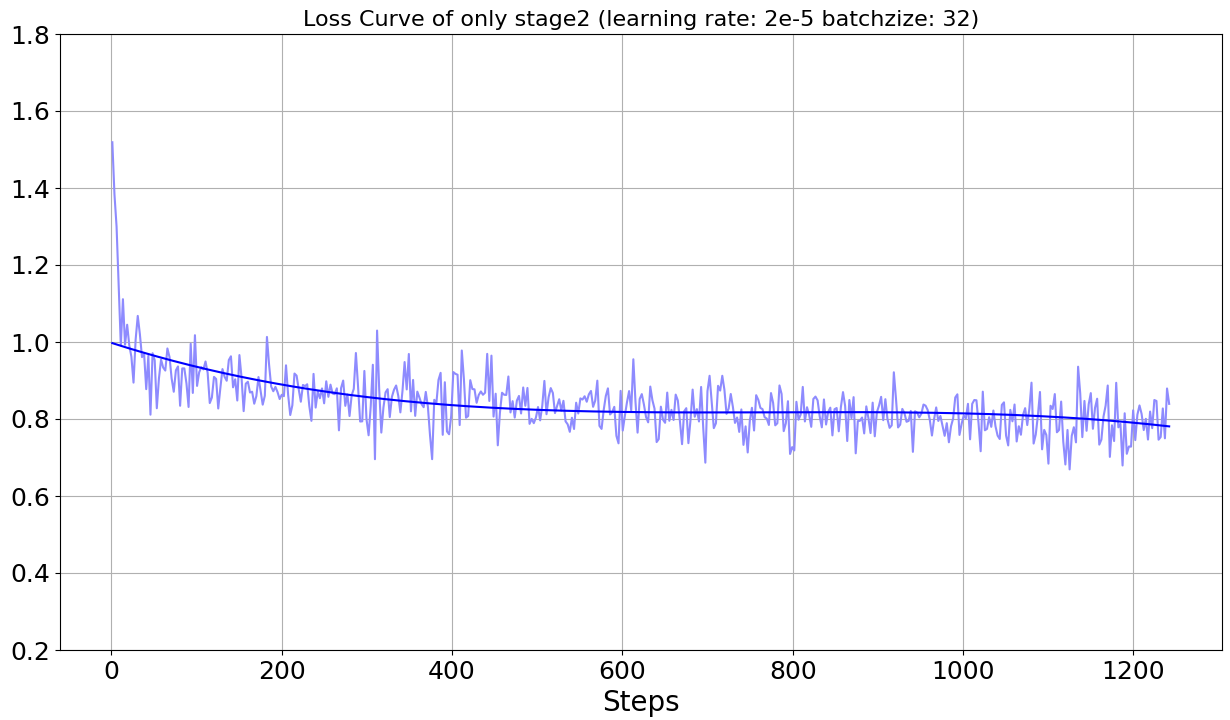} 
        \caption{VGA-7b-stage2}
    \end{minipage}
    \begin{minipage}{0.48\textwidth}
        \centering
        \includegraphics[width=\linewidth]{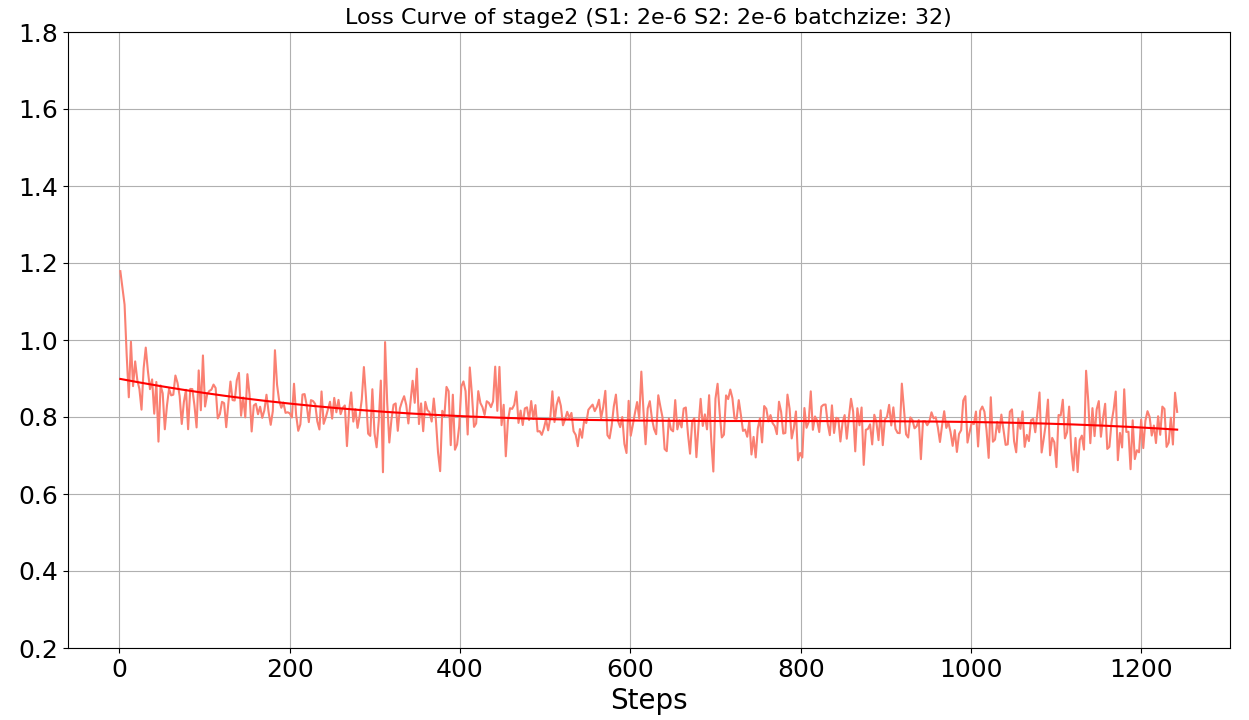} 
        \caption{VGA-7b-v1 (with FAC method)}
    \end{minipage}
    \begin{minipage}{0.7\textwidth}
        \centering
        \includegraphics[width=\linewidth]{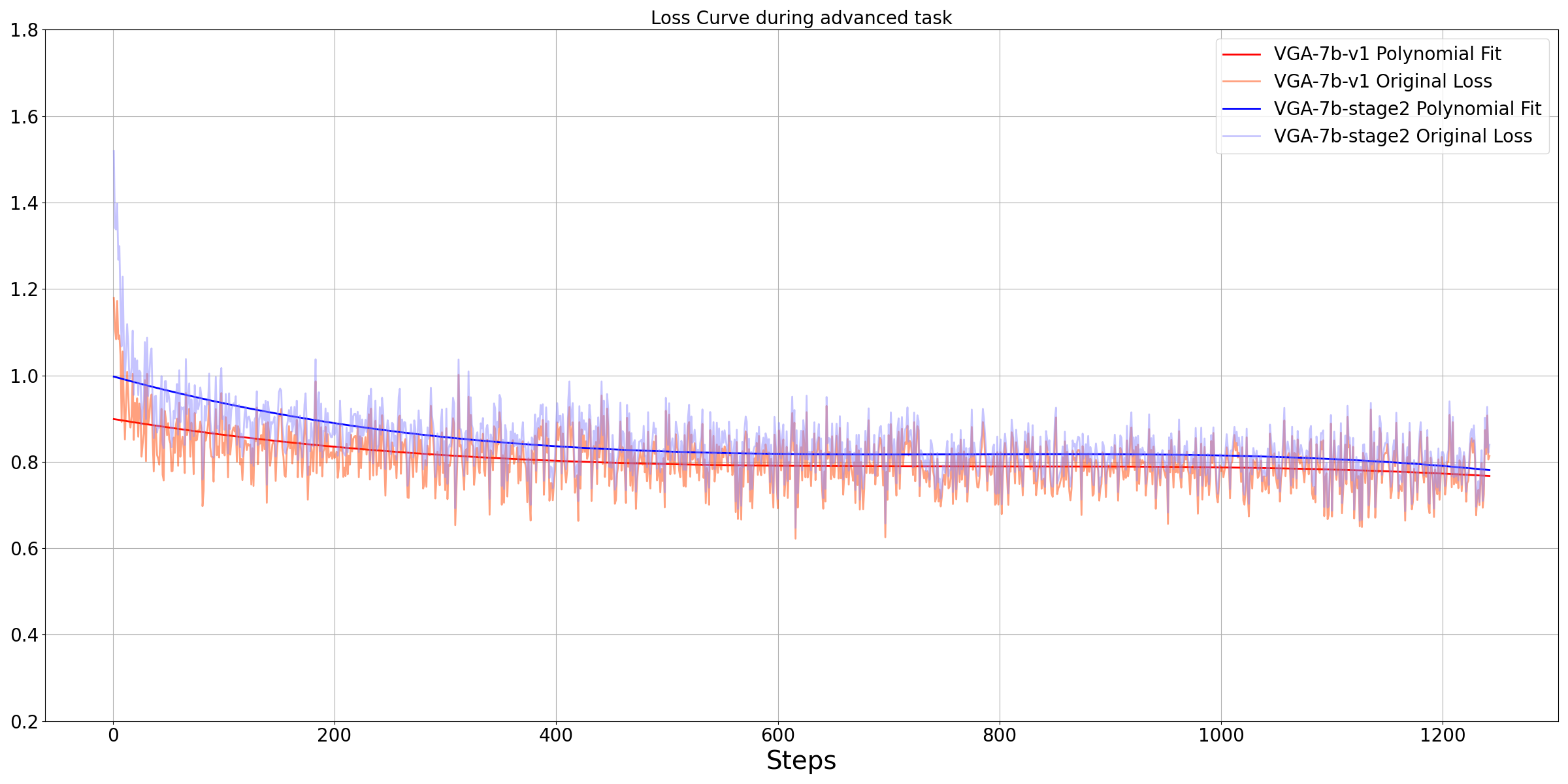} 
        \caption{Loss convergence comparing between VGA-7b-v1 and VGA-7b-stage2}
        \label{mean_att_l}
    \end{minipage}
\caption{}
\label{ablation}
\end{figure*}

We conduct ablation studies compare among base model (llava-v1.6-mistral-7b) and models fine-tuned by FAC method (VGA-7b-v1)and solely on advanced task. VGA-7b-v1 is fine-tuned based on model which pre-trained on foundation task, VGA-7b-stage2 is fine-tuned based on llava-v1.6-mistral-7b. As shown in Figure~\ref{ablation}, we compare their loss convergence during advanced task and result shows that the two models exhibit highly consistent loss trends, but VGA-7b-v1 demonstrates lower loss values. This indicates that our foundation task training approach, which involves freezing the question and response formats, does not degrade the model's intent-following performance. On the contrary, this training method allows the model to focus more on image information, leading to lower loss values during the advanced task training while maintaining similar trends.

We present some case in Table~\ref{tab:case4}, we evaluate this three models on real-world tasks. Compared to the baseline model llava-v1.6-mistral-7b, the responses of VGA-7b-stage2 incorporate the answer style of the reference method. Compared to VGA-7b-v1, the models that  not trained on foundation task shows limitations in GUI recognition accuracy, particularly in positional accuracy. This highlights the functionality and effectiveness of the two-phase training approach.

\subsection{Attention Analysis}
\label{appendix:attention}
To verify that our method increases the model's focus on extracting information from images when generating responses, we analyzed the attention values between image tokens and answer tokens. Specifically, we calculated the average attention contribution of each image token for every answer token during the generation process.. Additionally, we computed the total attention value of image tokens for each answer token during its generation. 

\begin{figure*}[h]  
    \centering
    \begin{minipage}{0.48\textwidth}
        \centering
        \includegraphics[width=\linewidth]{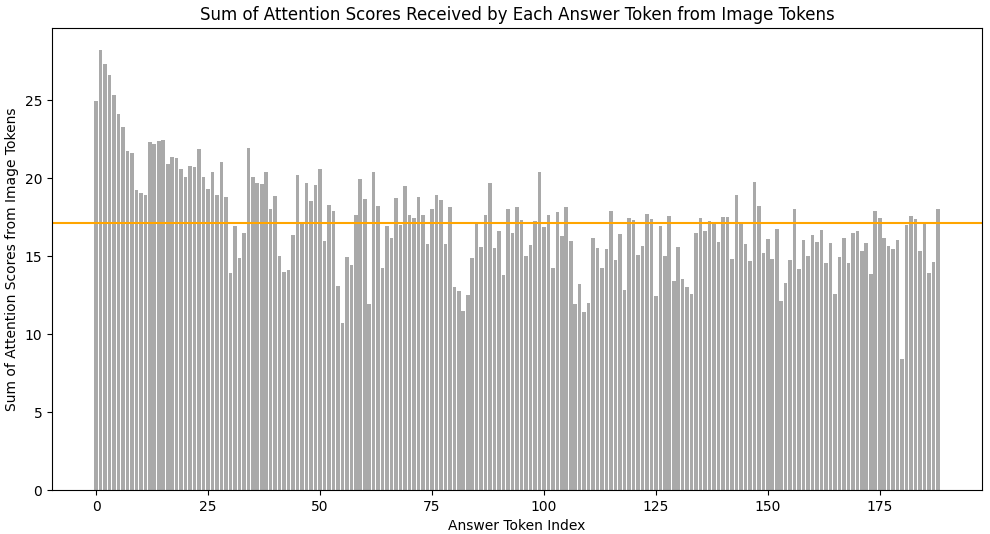} 
        \caption{llava-v1.6-mistral-7b}
    \end{minipage}
    \begin{minipage}{0.48\textwidth}
        \centering
        \includegraphics[width=\linewidth]{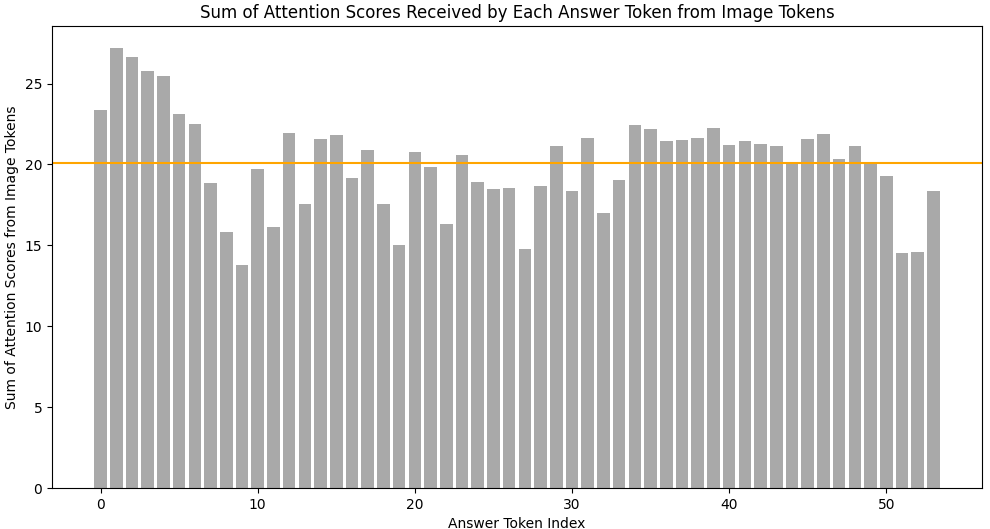} 
        \caption{VGA-7b-v1}
    \end{minipage}
    \begin{minipage}{0.48\textwidth}
        \centering
        \includegraphics[width=\linewidth]{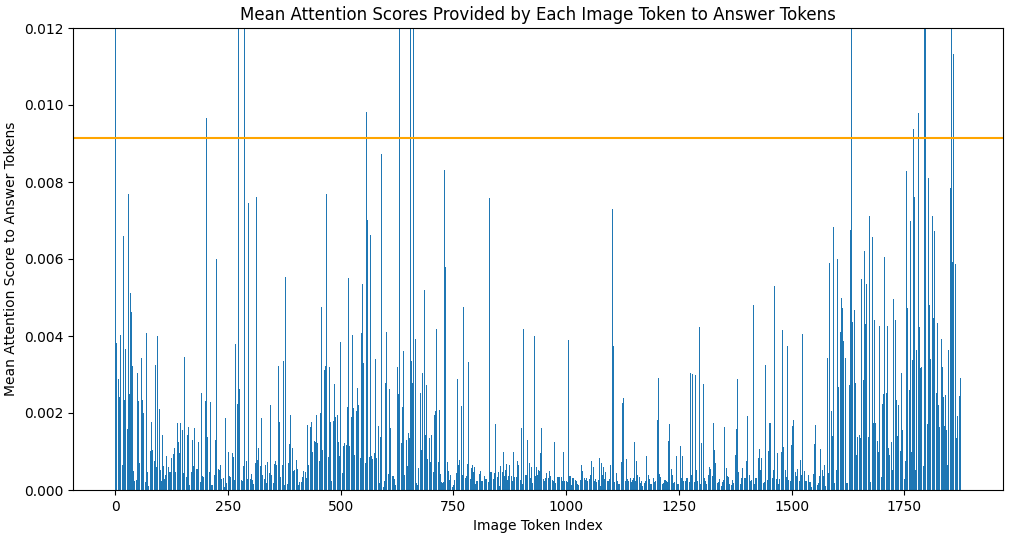} 
        \caption{llava-v1.6-mistral-7b}
    \end{minipage}
    \begin{minipage}{0.48\textwidth}
        \centering
        \includegraphics[width=\linewidth]{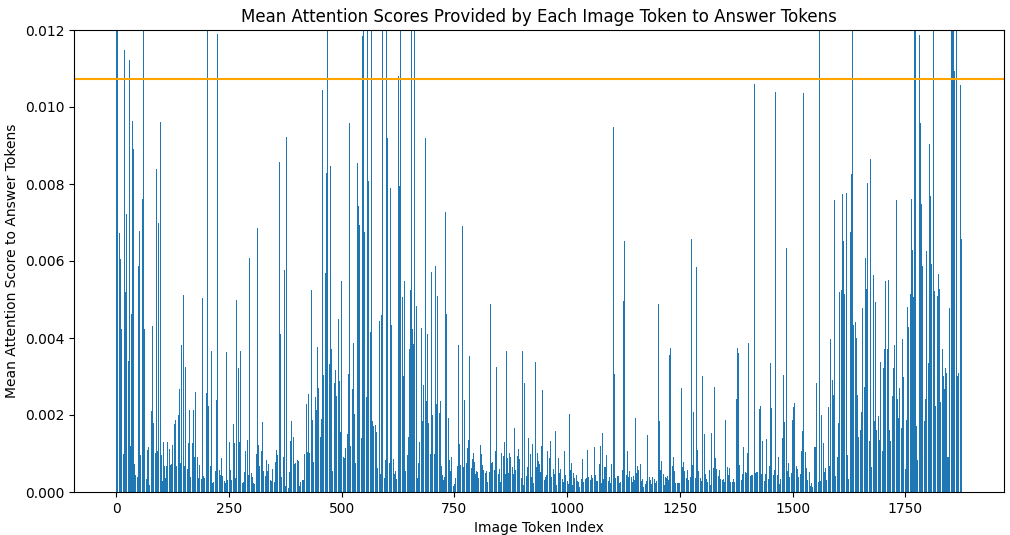} 
        \caption{VGA-7b-v1}
    \end{minipage}
    \begin{minipage}{0.48\textwidth}
        \centering
        \includegraphics[width=\linewidth]{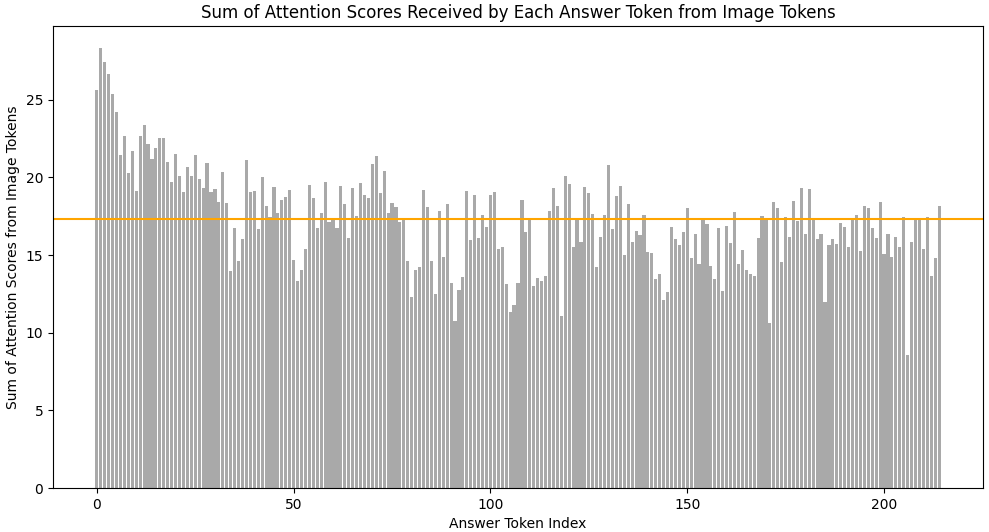} 
        \caption{llava-v1.6-mistral-7b}
    \end{minipage}
    \begin{minipage}{0.48\textwidth}
        \centering
        \includegraphics[width=\linewidth]{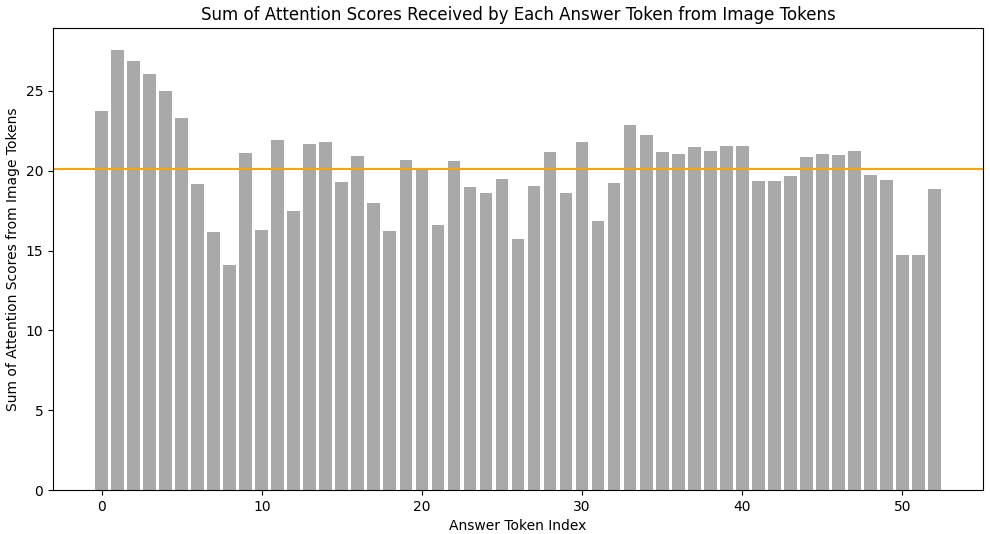} 
        \caption{VGA-7b-v1}
    \end{minipage}
    \begin{minipage}{0.48\textwidth}
        \centering
        \includegraphics[width=\linewidth]{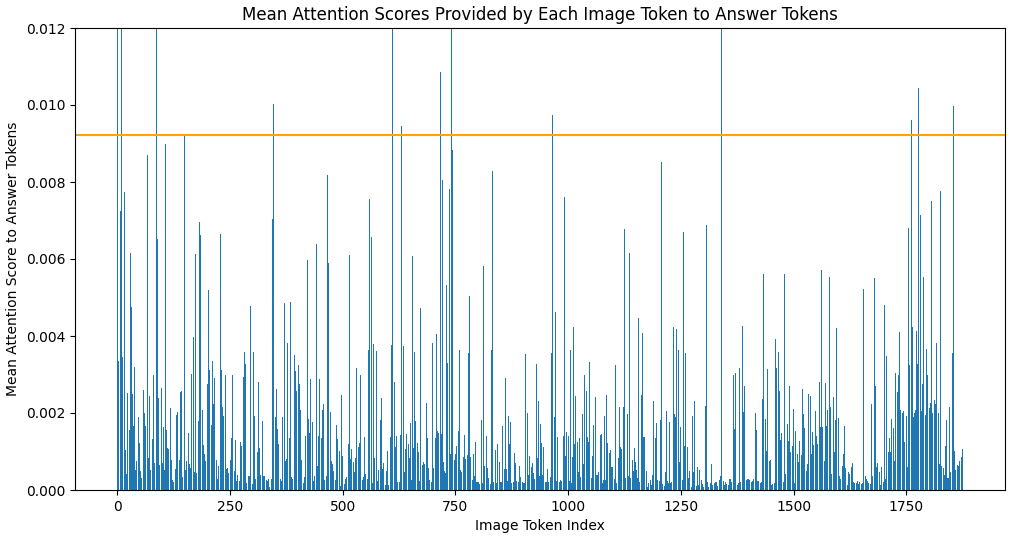} 
        \caption{llava-v1.6-mistral-7b}
    \end{minipage}
    \begin{minipage}{0.48\textwidth}
        \centering
        \includegraphics[width=\linewidth]{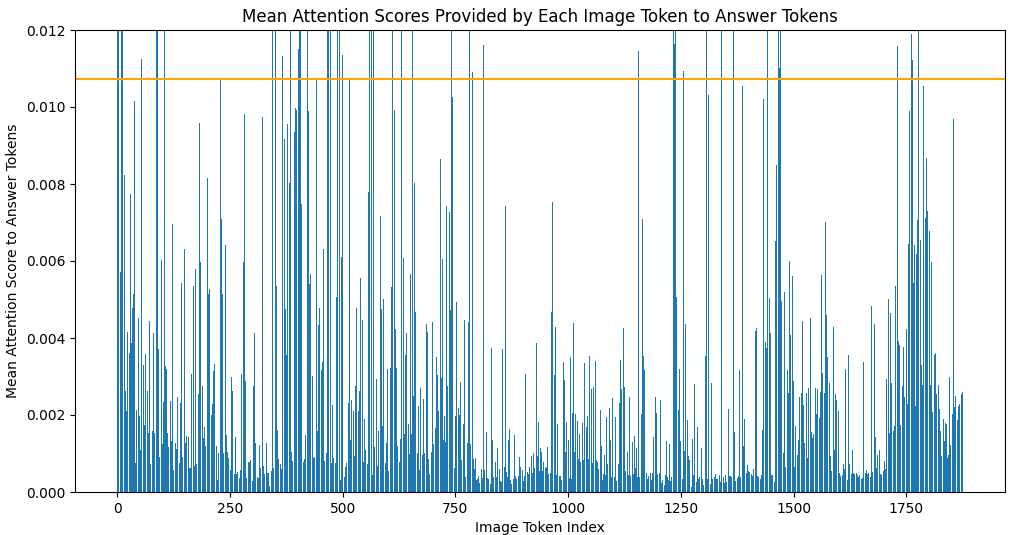} 
        \caption{VGA-7b-v1}
    \end{minipage}
\caption{}
\label{attention}
\end{figure*}

As shown in Figure~\ref{attention}, we compared our model VGA-7b-v1 with the baseline model llava-v1.6-mistral-7b using the same images and questions as input, and we recorded the differences in attention value. The results show that our model has significantly higher attention values to image tokens when generating responses compared to the llava model. This indicates that our model is more capable of capturing the content of the images.

\begin{table*}[h]
\centering
\begin{tabular}{p{2cm} c p{2cm} c p{2cm} c p{2cm} c p{2cm} c p{2cm} c p{2cm} c}

\hline
\fontsize{10}{14}\selectfont{\textbf{Experiment}}  & \fontsize{10}{14}\selectfont{\textbf{Data Size}} & \fontsize{10}{14}\selectfont{\textbf{Training Model}} & \fontsize{10}{14}\selectfont{\textbf{Learning Rate}} & \fontsize{10}{14}\selectfont{\textbf{Batch Size}} & \fontsize{10}{14}\selectfont{\textbf{GPUs}} & \fontsize{10}{14}\selectfont{\textbf{Time}} \\

\hline
\multicolumn{7}{c}{\fontsize{10}{14}\selectfont{\textbf{Foundation Task Training}}} \\
\hline
\fontsize{9}{10}\selectfont{Foundation Task} & \fontsize{9}{10}\selectfont{22.3k} &\fontsize{9}{10}\selectfont{Connector\& LLM} & \fontsize{9}{10}\selectfont{2e-6} & \fontsize{9}{10}\selectfont{16} & \fontsize{9}{10}\selectfont{1xA100 80G} & \fontsize{9}{10}\selectfont{16h}\\
\hline
\multicolumn{7}{c}{\fontsize{10}{14}\selectfont{\textbf{Advanced Comprehension Task Training}}}\\
\hline
\fontsize{9}{10}\selectfont{Advanced Task} & \fontsize{9}{10}\selectfont{41.4k}&\fontsize{9}{10}\selectfont{Connector\& LLM} & \fontsize{9}{10}\selectfont{2e-6} & \fontsize{9}{10}\selectfont{32} & \fontsize{9}{10}\selectfont{1xA100 80G} &\fontsize{9}{10}\selectfont{16h}\\
\hline
\end{tabular}
\caption{Hyperparameter in Training}
\label{training-parament}
\end{table*}

\begin{table*}[h]
    \centering
    \begin{tabular}{p{2.5cm}p{5cm}p{5cm}}
        \hline
        \textbf{\fontsize{12}{14}\selectfont Case in bench} & & \\
        \hline
        & \includegraphics[width=0.15\textwidth]{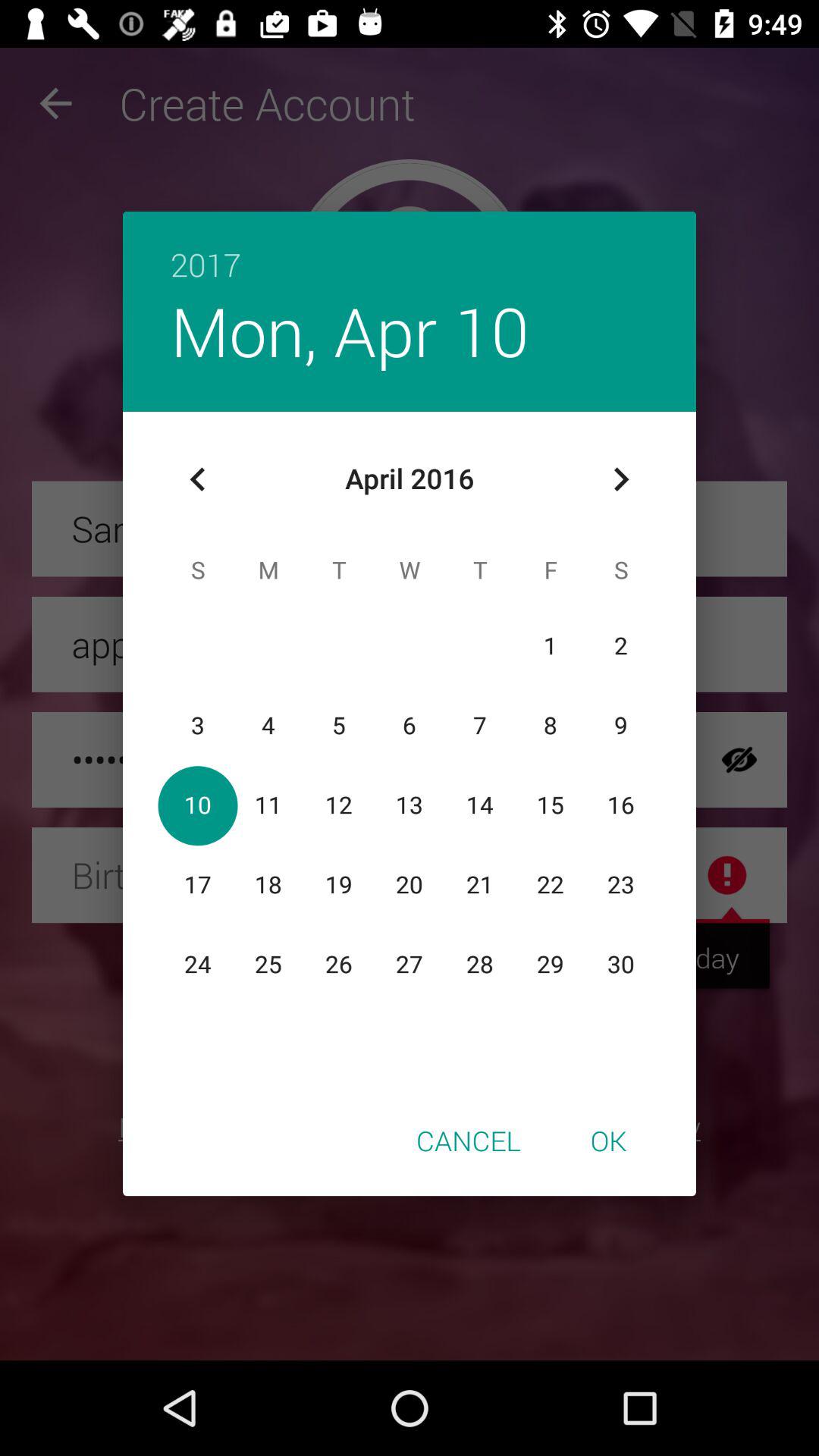} & \includegraphics[width=0.15\textwidth]{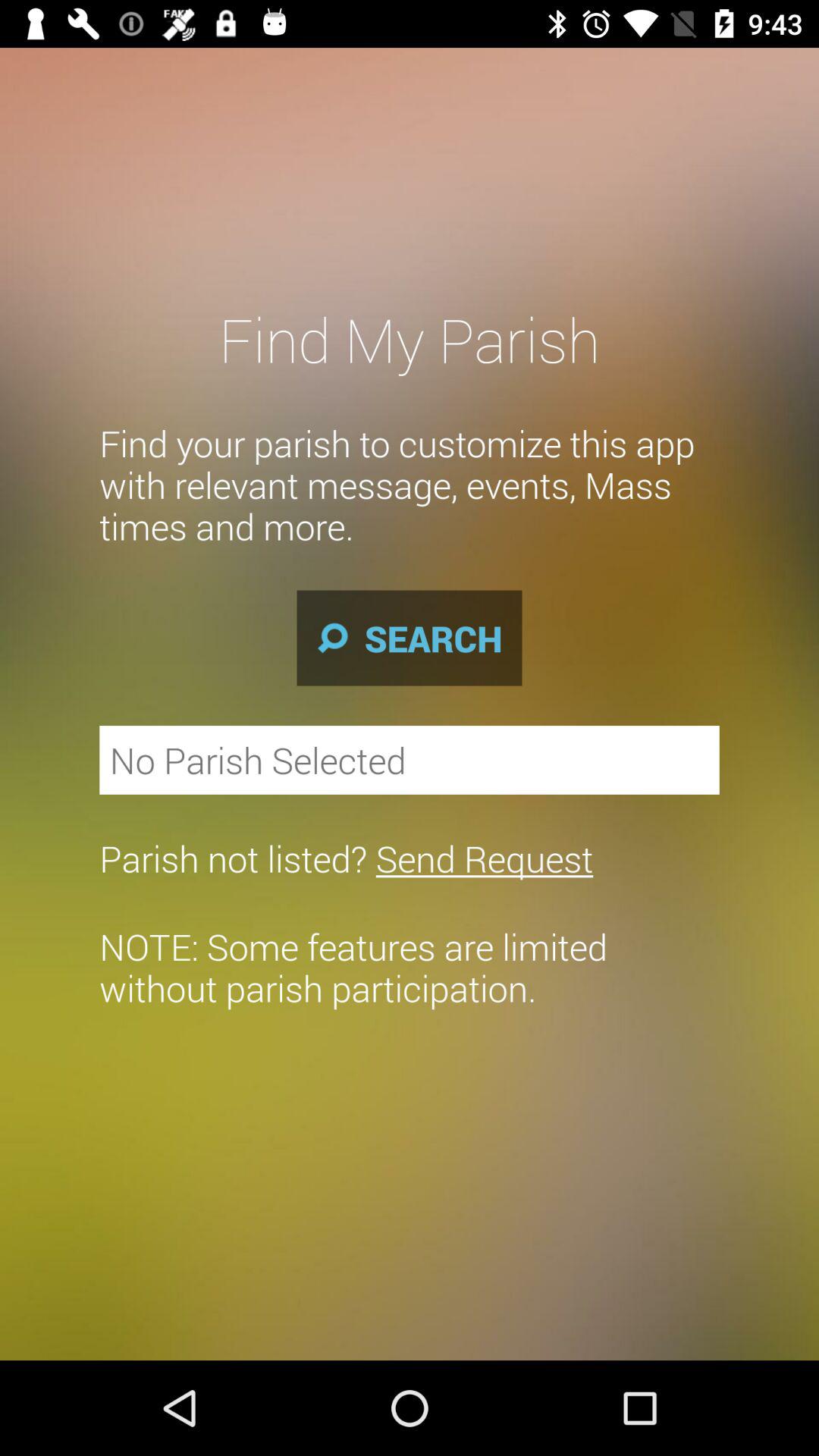}\\
        \fontsize{8}{10}\selectfont{User} & \fontsize{8}{10}\selectfont{\textcolor{blue}{What the day and date of the weekend before I selected in the middle of the page?}} & \fontsize{8}{10}\selectfont{\textcolor{blue}{How can I find my Parish?}} \\
        \fontsize{8}{10}\selectfont{VGA-7b-v1} & \fontsize{8}{10}\selectfont{ It is \textcolor{teal}{SAT, Apr 9, 2016} } & \fontsize{8}{10}\selectfont{ You can push \textcolor{teal}{"SEARCH"} button in the middle of the page with a \textcolor{teal}{blue} text. } \\
        \hline
        \fontsize{8}{10}\selectfont{llava-mistral-v1.6-7b} & \fontsize{8}{10}\selectfont{The image shows a calendar for April \textcolor{red}{2017}, and you have selected the date April 10, which is a Monday. The weekend before April 10 is Saturday, April 8, and Sunday, April 9. } & \fontsize{8}{10}\selectfont{ You can find your Parish by \textcolor{red}{customizing this app} with relevant message, events, Mass times and more. } \\
        \hline
        \fontsize{8}{10}\selectfont{Idefics2-8B} & \fontsize{8}{10}\selectfont{The day and date of the weekend before I selected are April 15th and 16th. \textcolor{red}{\textit{Year 2016 Missed.}}} & \fontsize{8}{10}\selectfont{To find your parish, you can use the \textcolor{teal}{search bar} provided on the app. This will allow you to input the name of your parish or search for it using other relevant information.} \\
        \hline
        \fontsize{8}{10}\selectfont{MiniCPM-V-2\_5} & \fontsize{8}{10}\selectfont{ The day and date of the weekend before April 10, which is a Monday, would be Sunday, April 9. \textcolor{red}{\textit{Year 2016 Missed.} }} & \fontsize{8}{10}\selectfont{To find your Parish using the app, you can \textcolor{red}{customize it with relevant messages, events, and Mass times by selecting a parish from the list provided.} } \\
        \hline
    \end{tabular}
    \caption{Case in the GUI-benchmark}
    \label{tab:case1}
    
\end{table*}

\begin{table*}[h]
    \centering
    \begin{tabular}{p{2.5cm}p{5cm}p{5cm}}
        \hline
        \textbf{\fontsize{12}{14}\selectfont Case in bench} & & \\
        \hline
        & \includegraphics[width=0.15\textwidth]{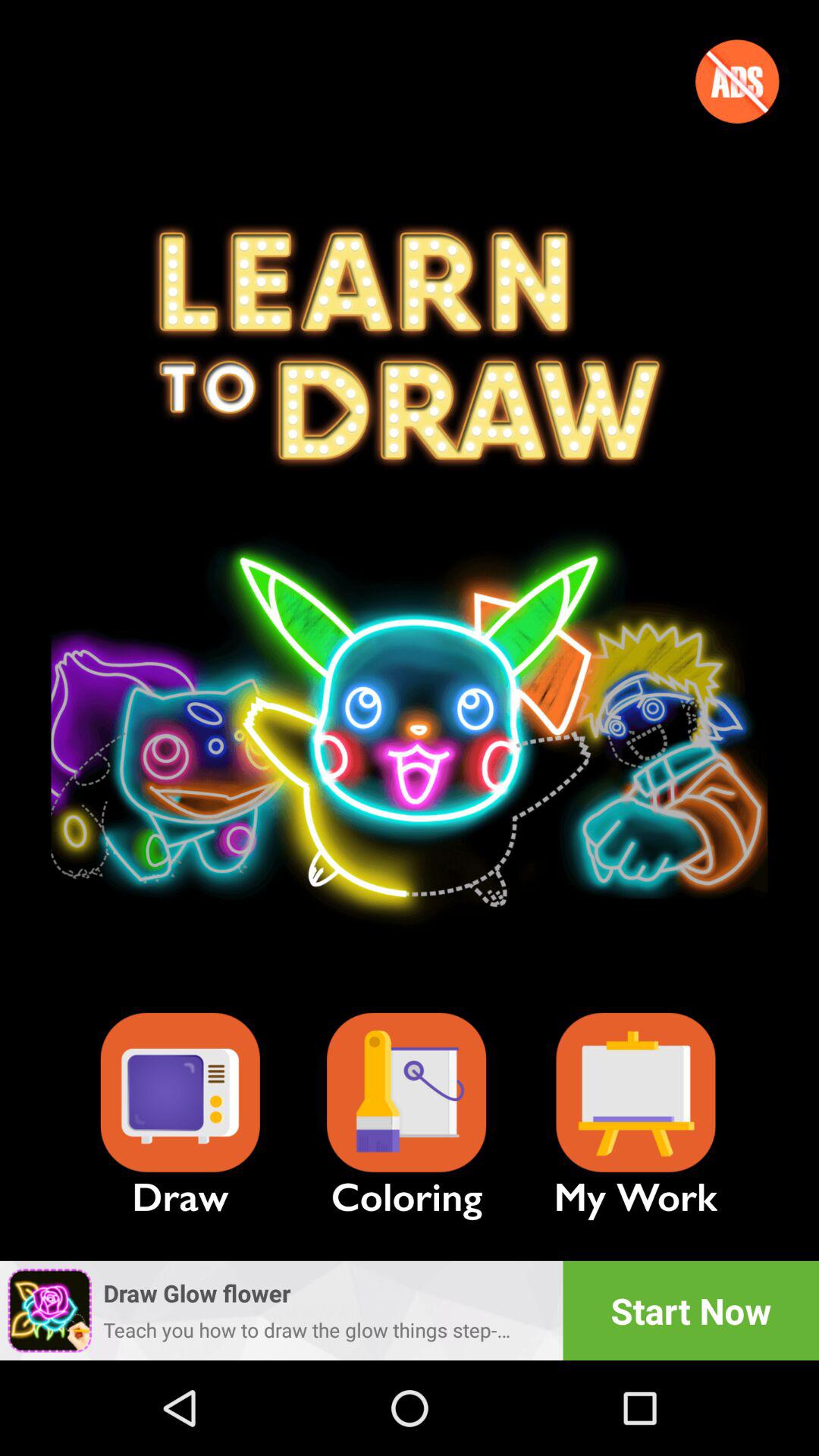} & \includegraphics[width=0.15\textwidth]{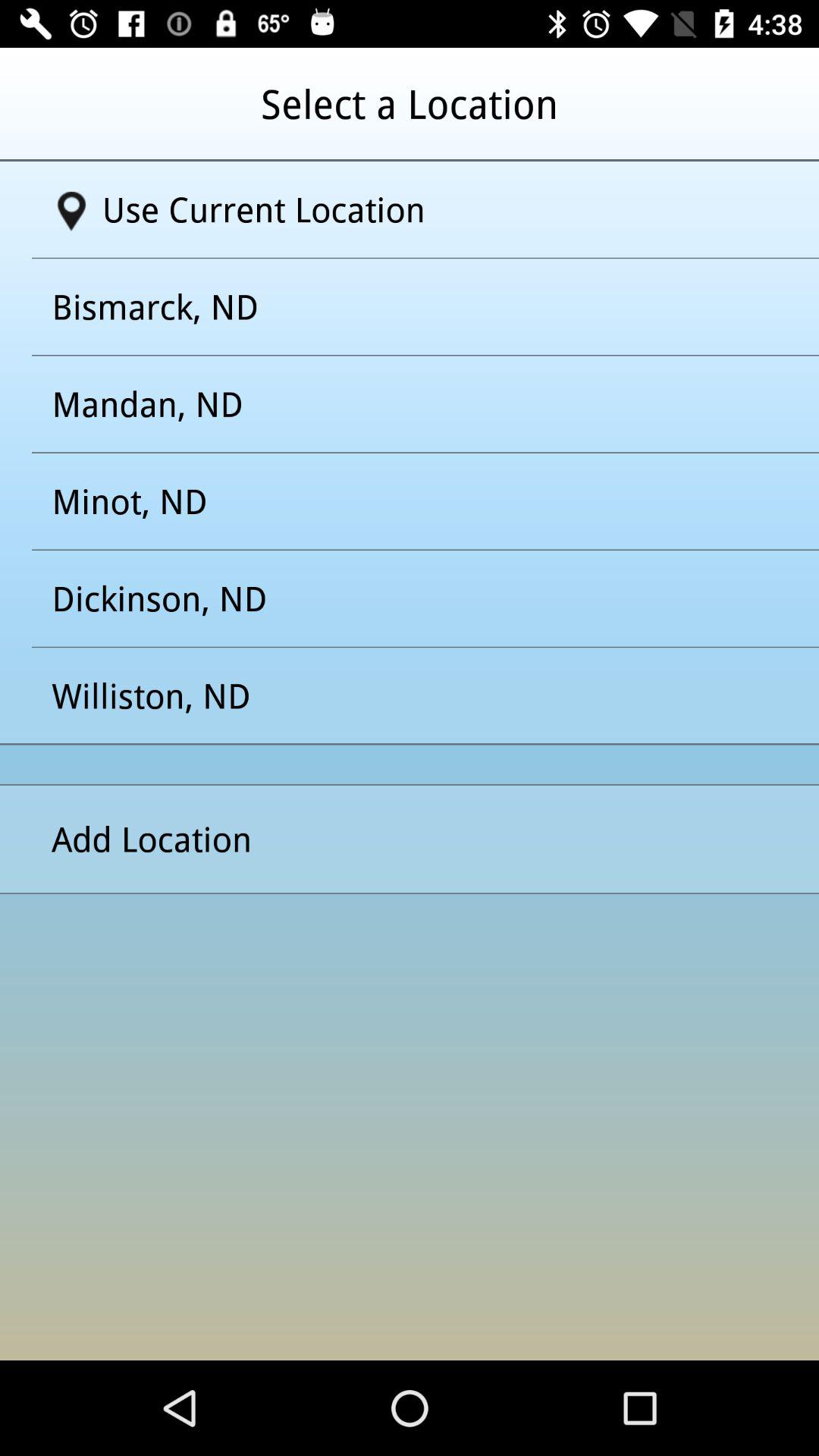}\\
        \fontsize{8}{10}\selectfont{User} & \fontsize{8}{10}\selectfont{\textcolor{blue}{How can I start Drawing?}} & \fontsize{8}{10}\selectfont{\textcolor{blue}{Am I use current location right now?}} \\
        \fontsize{8}{10}\selectfont{VGA-7b-v1} & \fontsize{8}{10}\selectfont{ You can click the \textcolor{teal}{orange "Draw" button in the bottom left corner of the page.} } & \fontsize{8}{10}\selectfont{ \textcolor{teal}{Yes}, you are using you current location. The location icon is in the \textcolor{teal}{first line named "Use Current Location".} } \\
        \hline
        \fontsize{8}{10}\selectfont{llava-mistral-v1.6-7b} & \fontsize{8}{10}\selectfont{ To start drawing, you can use the \textcolor{red}{Draw app} on your device, which is available in the app store. } & \fontsize{8}{10}\selectfont{ \textcolor{teal}{Yes}, you are using your current location right now. } \\
        \hline
        \fontsize{8}{10}\selectfont{Idefics2-8B} & \fontsize{8}{10}\selectfont{ To start drawing, you can either \textcolor{teal}{tap on the "Draw" button} or \textcolor{red}{select one of the three options provided (Coloring, My Work, or Start Now).} } & \fontsize{8}{10}\selectfont{\textcolor{teal}{Yes}, I am using your current location to provide information about nearby places of interest.} \\
        \hline
        \fontsize{8}{10}\selectfont{MiniCPM-V-2\_5} & \fontsize{8}{10}\selectfont{ You can click the \textcolor{teal}{"Draw" button.} } & \fontsize{8}{10}\selectfont{ \textcolor{red}{No, you are not currently using your current location.} } \\
        \hline
    \end{tabular}
    \caption{Case in the GUI-benchmark}
    \label{tab:case2}
    
\end{table*}

\begin{table*}[h]
    \centering
    \begin{tabular}{p{2.5cm}p{5cm}p{5cm}}
        \hline
        \textbf{\fontsize{12}{14}\selectfont Case in bench} & & \\
        \hline
        & \includegraphics[width=0.15\textwidth]{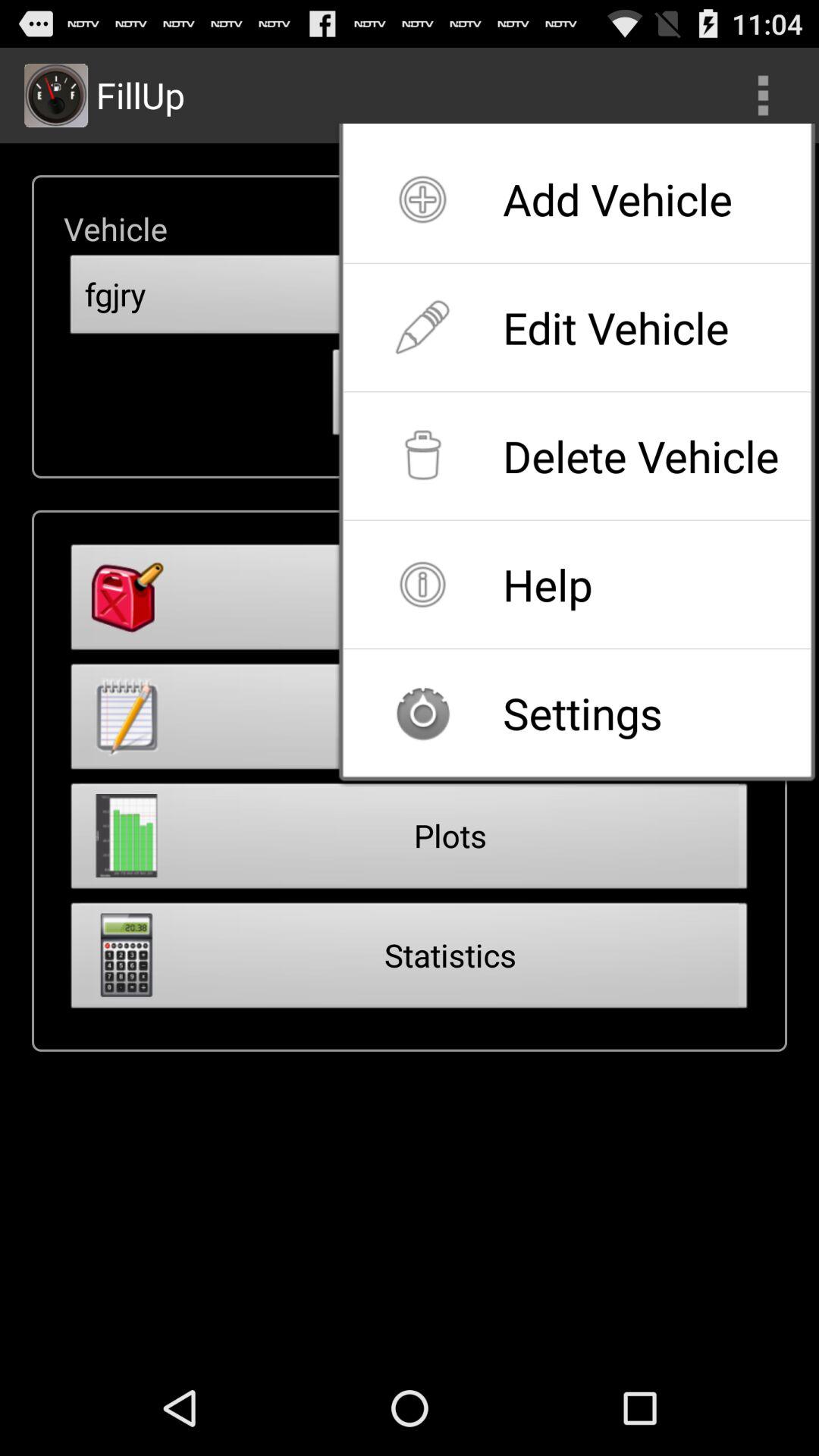} & 
        \includegraphics[width=0.15\textwidth]{0072141.jpg}\\
        \fontsize{8}{10}\selectfont{User} & 
        \fontsize{8}{10}\selectfont{\textcolor{blue}{How can I remove the vehicle?}} &
        \fontsize{8}{10}\selectfont{\textcolor{blue}{Where can I find the work I drew?}} \\
        \fontsize{8}{10}\selectfont{VGA-7b-v1} &
        \fontsize{8}{10}\selectfont{You can remove the vehicle by clicking on the 'Delete Vehicle' option \textcolor{orange}{in the dropdown menu at the top right of the page. It's the third option from the top.} } &
        \fontsize{8}{10}\selectfont{ You can find the work you've drawn by clicking on the \textcolor{orange}{orange square icon labeled 'My Work' at the bottom right of the page.} } \\
        \hline
        \fontsize{8}{10}\selectfont{GPT-4o} &
        \fontsize{8}{10}\selectfont{ To remove the vehicle, tap on "Delete Vehicle" from \textcolor{orange}{the menu shown.} } &
        \fontsize{8}{10}\selectfont{You can find the paintings you drew by clicking on the "My Work" button. } \\
        \hline
        \fontsize{8}{10}\selectfont{GPT-4V} &
        \fontsize{8}{10}\selectfont{ You can remove the vehicle by selecting the "Delete Vehicle" option \textcolor{orange}{in the menu shown on the app.}} &
        \fontsize{8}{10}\selectfont{You can find the paintings you drew in the "My Work" section of the app shown in the image.} \\
        \hline
        \fontsize{8}{10}\selectfont{MiniCPM-V-2\_5} &
        \fontsize{8}{10}\selectfont{To remove a vehicle from the application, you would typically select the 'Delete Vehicle' option \textcolor{orange}{from the menu.} } &
        \fontsize{8}{10}\selectfont{The "My Work" icon suggests a section where users can view their past drawings. } \\
        \hline
    \end{tabular}
    \caption{Case in the GUI-benchmark}
    \label{tab:case3}
    
\end{table*}

\begin{table*}[h]
    \centering
    \begin{tabular}{p{2.5cm}p{5cm}p{5cm}}
        \hline
        \textbf{\fontsize{12}{14}\selectfont{Ablation Study} } & & \\
        \hline
        & \includegraphics[width=0.15\textwidth]{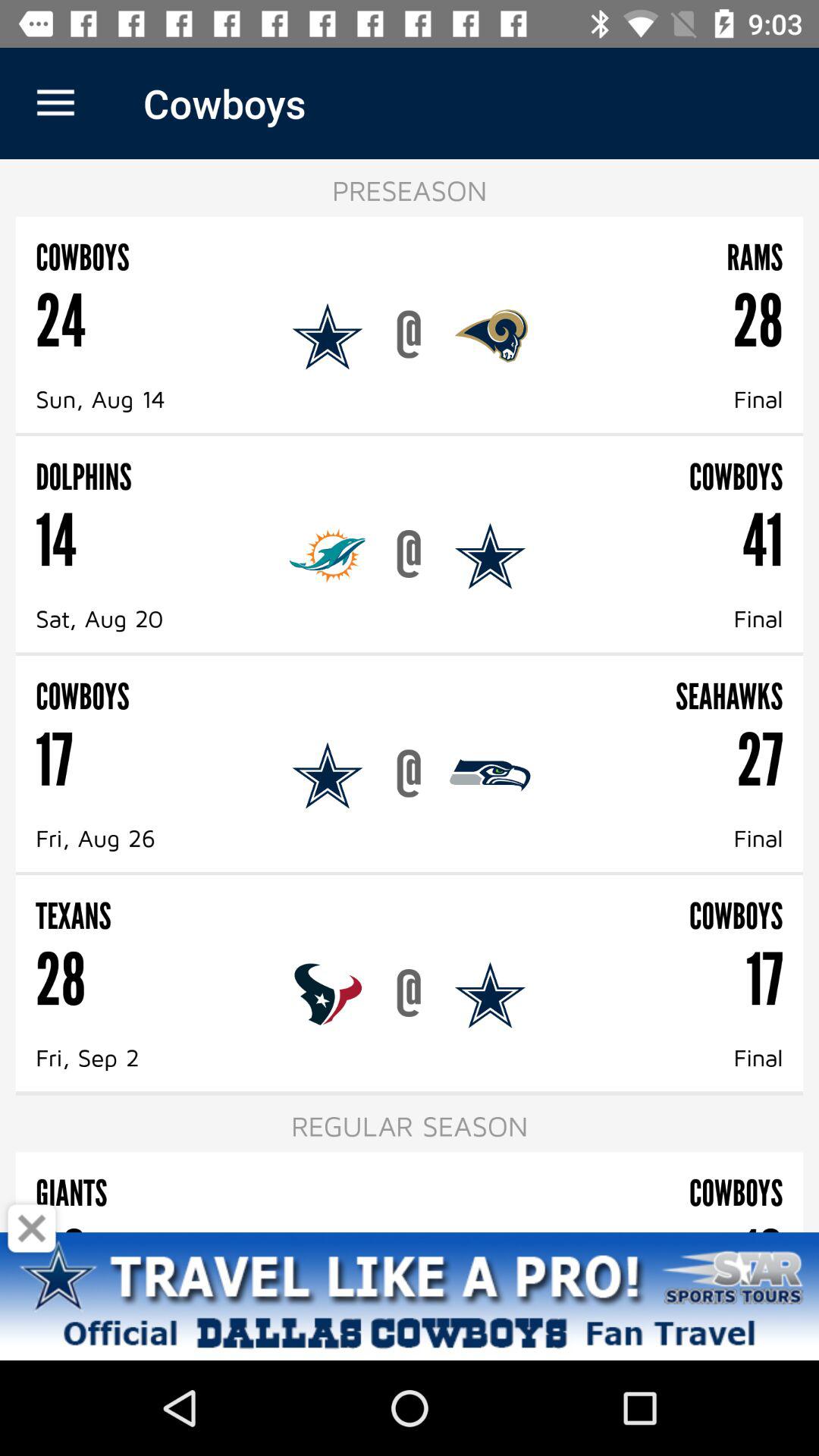} & 
        \includegraphics[width=0.15\textwidth]{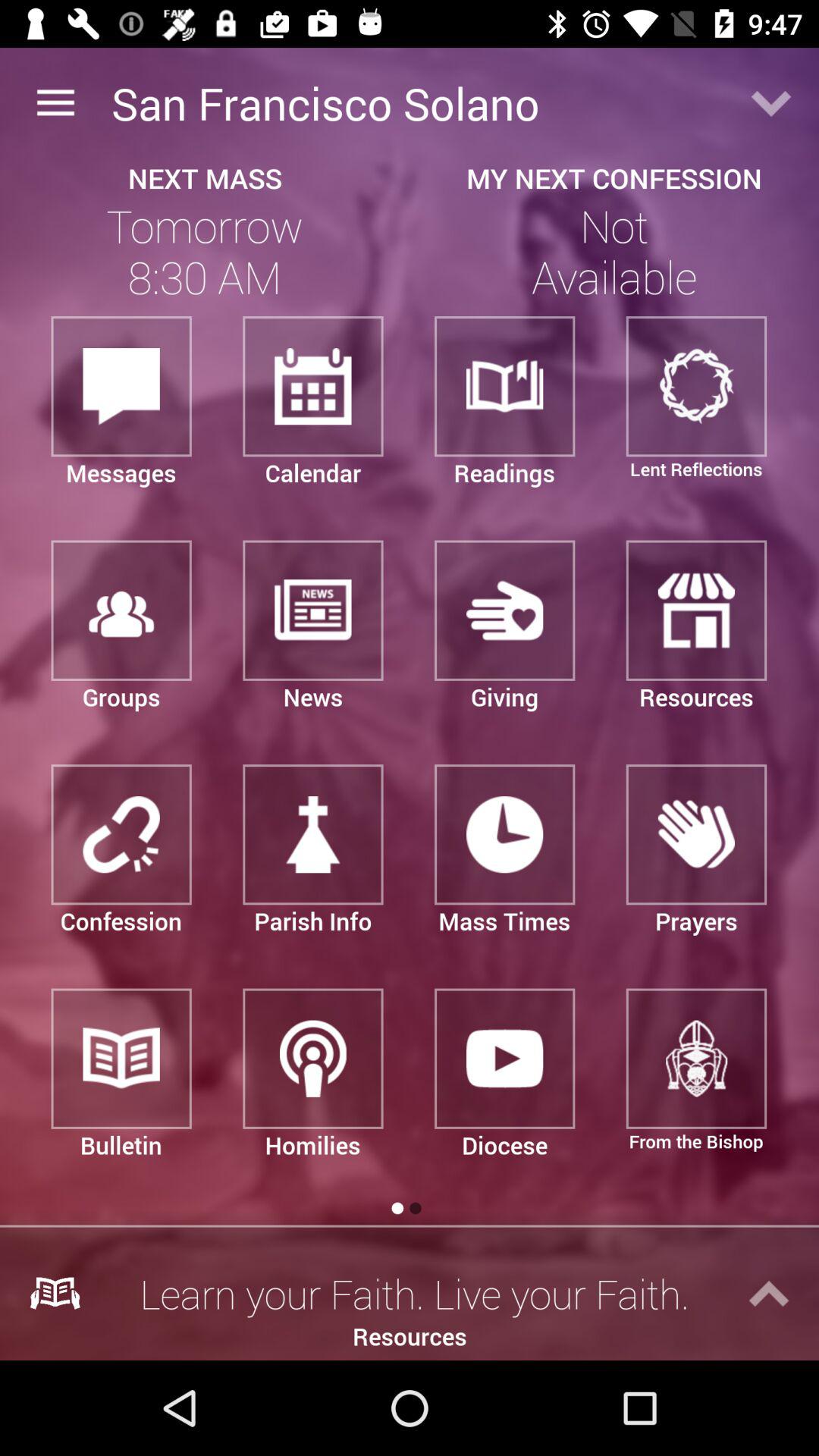}\\
        \fontsize{8}{10}\selectfont{User} & 
        \fontsize{8}{10}\selectfont{\textcolor{blue}{Who will play against each other on Sun, Aug 14?}} &
        \fontsize{8}{10}\selectfont{\textcolor{blue}{where is the diocese?}} \\
        \fontsize{8}{10}\selectfont{VGA-7b-v1} &
        \fontsize{8}{10}\selectfont{The \textcolor{teal}{Cowboys and the Rams} will play against each other on sun, aug 14.  } &
        \fontsize{8}{10}\selectfont{The Diocese icon is \textcolor{teal}{in the bottom row, second from the left}. It is a white icon with a video camera symbol.} \\
        \hline
        \fontsize{8}{10}\selectfont{VGA-7b-stage2} &
        \fontsize{8}{10}\selectfont{ The \textcolor{red}{Cowboys will play against the Dolphins} on Sun, Aug 14..} &
        \fontsize{8}{10}\selectfont{The Diocese is \textcolor{red}{in the second row, second column}. } \\
        \hline
        \fontsize{8}{10}\selectfont{llava-v1.6-mistral-7b} &
        \fontsize{8}{10}\selectfont{ \textcolor{red}{Cowboys and dolphins}.} &
        \fontsize{8}{10}\selectfont{The diocese icon is located \textcolor{red}{in the bottom left corner} of the app.} \\
        \hline
    \end{tabular}
    \caption{Ablation study of the model trained with foundation task and without. llava-v1.6-mistral-7b is the base model, VGA-7B-stage2 is the model trained solely on advanced task.}
    \label{tab:case4}
    
\end{table*}

\begin{table*}[h]
    \centering
    \begin{tabular}{p{2.5cm}p{5cm}p{5cm}}
        \hline
        \textbf{\fontsize{12}{14}\selectfont{Real world} } & & \\
        \hline
        & \includegraphics[width=0.15\textwidth]{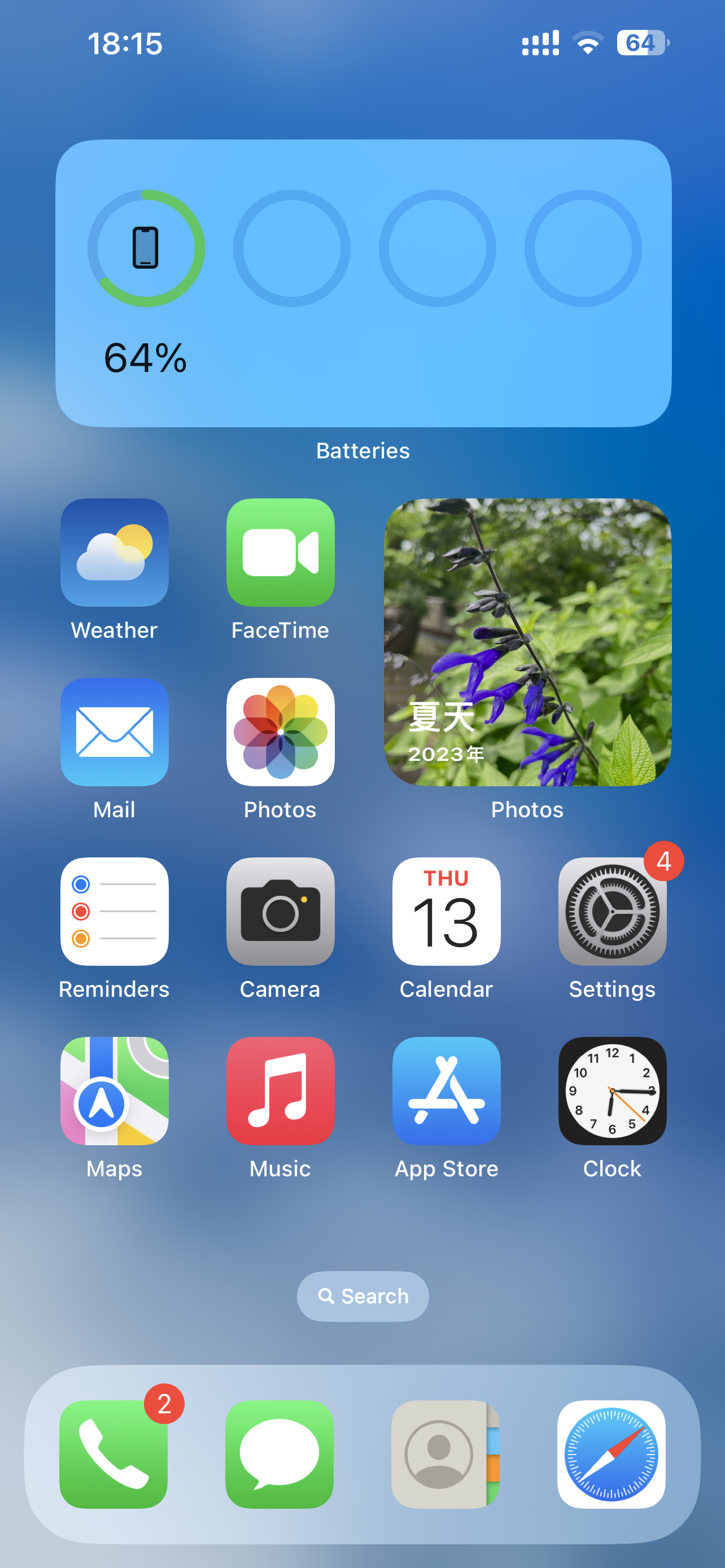} & 
        \includegraphics[width=0.15\textwidth]{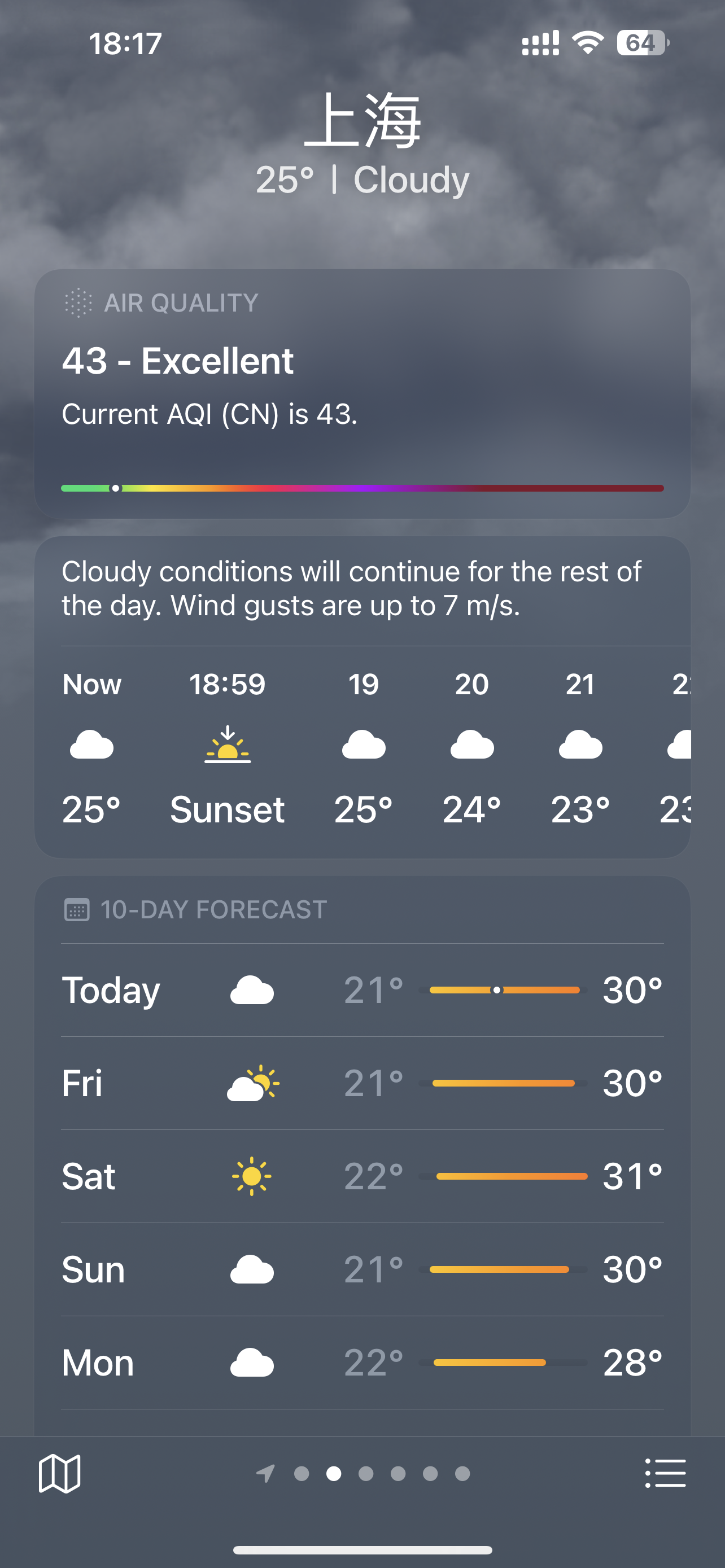}\\
        \fontsize{8}{10}\selectfont{User} & 
        \fontsize{8}{10}\selectfont{\textcolor{blue}{Please open the weather app, click?}} &
        \fontsize{8}{10}\selectfont{\textcolor{blue}{What is the weather on Sat?}} \\
        \fontsize{8}{10}\selectfont{VGA-7b-v1} &
        \fontsize{8}{10}\selectfont{<0.12, 0.28> } &
        \fontsize{8}{10}\selectfont{The weather in Sat is sunny with a temperature of 22°C.} \\

        \fontsize{8}{10}\selectfont{User} & 
        \fontsize{8}{10}\selectfont{\textcolor{blue}{Has anyone called me?}} &
        \fontsize{8}{10}\selectfont{\textcolor{blue}{What is the highest temperature today?}} \\
        \fontsize{8}{10}\selectfont{VGA-7b-v1} &
        \fontsize{8}{10}\selectfont{Yes, there is a call icon with a red notification badge on it.} &
        \fontsize{8}{10}\selectfont{The highest temperature today is 30°C. You can find it in the bottom section of the page, under the `10-DAY FORECAST' section.} \\

        \fontsize{8}{10}\selectfont{User} & 
        \fontsize{8}{10}\selectfont{\textcolor{blue}{Could you help me to play some song?}} &
        \fontsize{8}{10}\selectfont{\textcolor{blue}{what is the temperature in 19:00?}} \\
        \fontsize{8}{10}\selectfont{VGA-7b-v1} &
        \fontsize{8}{10}\selectfont{To play a song, tap on the icon that looks like a music note. It's located in the bottom row, second from the left.} &
        \fontsize{8}{10}\selectfont{The temperature in 19:00 is 25°C.} \\

        \fontsize{8}{10}\selectfont{User} & 
        \fontsize{8}{10}\selectfont{\textcolor{blue}{What is date of the week today?}} &
        \fontsize{8}{10}\selectfont{\textcolor{blue}{What is the weather like now?}} \\
        \fontsize{8}{10}\selectfont{VGA-7b-v1} &
        \fontsize{8}{10}\selectfont{Today's date of the week is Thursday.} &
        \fontsize{8}{10}\selectfont{The current weather is 25°C and cloudy. } \\

        \fontsize{8}{10}\selectfont{User} & 
        \fontsize{8}{10}\selectfont{\textcolor{blue}{Open the camera.}} &
        \fontsize{8}{10}\selectfont{\textcolor{blue}{How will the weather change next?}} \\
        \fontsize{8}{10}\selectfont{VGA-7b-v1} &
        \fontsize{8}{10}\selectfont{To open the camera, tap on the screen at <0.34, 0.7>. }&
        \fontsize{8}{10}\selectfont{The weather will continue for the rest of the day. Wind gusts are up to 7 m/s. }\\
        \fontsize{8}{10}\selectfont{User} & 
        \fontsize{8}{10}\selectfont{\textcolor{blue}{What is the setting bounds?}} &
        \fontsize{8}{10}\selectfont{\textcolor{blue}{How about the air quality?}} \\
        \fontsize{8}{10}\selectfont{VGA-7b-v1} &
        \fontsize{8}{10}\selectfont{The setting bounds are <0.762, 0.566, 0.922, 0.692>.} &
        \fontsize{8}{10}\selectfont{The air quality is currently labeled as 'Excellent' with a value of 43.} \\
        \hline
        \hline
    \end{tabular}
    \caption{Case in the real world}
    \label{tab:case5}
    
\end{table*}

\end{document}